\documentclass[conference]{IEEEtran}
\IEEEoverridecommandlockouts

\usepackage{booktabs} 

\usepackage{amsmath}
\usepackage{amssymb}
\usepackage{graphicx}

\usepackage{here}
\usepackage{url}
\usepackage{comment}
\usepackage{algorithm}
\usepackage{algorithmicx}
\usepackage[noend]{algpseudocode}

\usepackage{subfigure}

\newtheorem{thm}{Theorem}
\newcommand{\av}{{\bf a}}
\newcommand{\x}{{\bf x}}
\newcommand{\y}{{\bf y}}
\newcommand{\z}{{\bf z}}
\newcommand{\R}{{\bf R}}
\newcommand{\N}{{\bf N}}
\newcommand{\E}{{\bf E}}
\newcommand{\D}{\mathcal{D}}

\usepackage{cite}
\usepackage{amsmath,amssymb,amsfonts}
\usepackage{textcomp}
\usepackage{xcolor}
\def\BibTeX{{\rm B\kern-.05em{\sc i\kern-.025em b}\kern-.08em
    T\kern-.1667em\lower.7ex\hbox{E}\kern-.125emX}}
\begin{document}

\title{Space-efficient Feature Maps for \\
String Alignment Kernels}



\author{\IEEEauthorblockN{Yasuo Tabei}
\IEEEauthorblockA{RIKEN Center for \\ Advanced Intelligence Project \\
yasuo.tabei@riken.jp}
\and
\IEEEauthorblockN{Yoshihiro Yamanishi}
\IEEEauthorblockA{Kyushu Institute of Technology \\
yamani@bio.kyutech.ac.jp}
\and
\IEEEauthorblockN{Rasmus Pagh}
\IEEEauthorblockA{BARC and \\ IT University of Copenhagen \\
pagh@itu.dk}
}

\maketitle

\begin{abstract}
String kernels are attractive data analysis tools for analyzing string data. Among them, alignment kernels are known for their high prediction accuracies in
string classifications when tested in combination with SVM in various applications.
However, alignment kernels have a crucial drawback in that they scale poorly due to their quadratic computation complexity in the number of input strings, which limits large-scale applications in practice.
We address this need by presenting the first approximation for string alignment kernels,
which we call {\em space-efficient feature maps for edit distance with moves (SFMEDM)}, by leveraging a metric embedding named {\em edit sensitive parsing~(ESP)} and {\em feature maps (FMs)} of {\em random Fourier features (RFFs)} for large-scale string analyses.
The original FMs for RFFs consume a huge amount of memory proportional to the dimension $d$ of input vectors and the dimension $D$ of output vectors,
which prohibits its large-scale applications.
We present novel {\em space-efficient feature maps (SFMs)} of RFFs for a space reduction from $O(dD)$ of the original FMs to $O(d)$ of SFMs with a theoretical
guarantee with respect to concentration bounds.
We experimentally test SFMEDM on its ability to learn SVM for large-scale string classifications with various massive string data, and we demonstrate the superior performance of SFMEDM with respect to prediction accuracy, scalability and computation efficiency.
\end{abstract}


\section{Introduction}
Massive string data are now ubiquitous throughout research and industry, in areas such as biology, chemistry, natural language processing and data science.
For example, e-commerce companies face a serious problem in analyzing huge datasets of user reviews, question answers and purchasing histories~\cite{He16,McAuley15}.
In biology, homology detection from huge collections of protein and DNA sequences is an important part for their functional analyses~\cite{Saigo04}.
There is therefore a strong need to develop powerful methods to make best use of massive string data on a large-scale.

Kernel methods~\cite{Hofmann08} are attractive data analysis tools
because they can approximate any (possibly non-linear) function or decision boundary well with enough training data.
In kernel methods, a kernel matrix a.k.a.~Gram matrix is computed from training data and {\em non-linear support vector machines (SVM)} are trained on the matrix.
Although it is known that kernel methods achieve high prediction accuracy for various tasks such as classification and regression,
they scale poorly due to a quadratic complexity in the number of training data~\cite{Joachims06,Ferris03}.
In addition, calculation of a classification requires, in the worst case, linear time in the number of training data,
which limits large-scale applications of kernel methods in practice.

String kernels~\cite{Gartner03} are kernel functions that operate on strings, and a variety of string kernels using string similarity measures have been proposed~\cite{Leslie02, Saigo04, Cuturi11, Lodhi02}. 
As state-of-the-art string kernels, string alignment kernels are known for high prediction accuracy in 
string classifications, such as remote homology detection for protein sequences~\cite{Saigo04} and time-series classifications~\cite{Zhou10, Cuturi11}, when tested in combination with SVM.
However, alignment kernels have a crucial drawback; that is, as in other kernel methods, they scale poorly due to their quadratic computation complexity in the number of training data.

Kernel approximations using {\em feature maps~(FMs)} have been proposed to 
solve the scalability issues regarding kernel methods.
FMs project training data into low-dimensional vectors such that the kernel value (similarity) between each pair of training data is approximately equal to 
the inner product of the corresponding pair of low dimensional vectors. 
Then, {\em linear SVM} are trained on the projected vectors, thereby significantly improving the scalability, while preserving their prediction accuracy. 
Although a variety of kernel approximations using FMs for enhancing the scalability of kernel methods have been proposed (e.g., Jaccard kernels~\cite{Li11}, 
polynomial kernels~\cite{Pham13} and Min-Max kernels~\cite{Li16}), and 
{\em random Fourier features~(RFFs)}~\cite{Rahimi07} are an approximation of shift-invariant kernels (e.g., Laplacian and radial basis function~(RBF) kernels), 
approximation for string alignment kernels has not been studied.
Thus, an important open challenge, which is required for large-scale analyses of string data, is to develop a kernel approximation for string alignment kernels.

Several metric embeddings for string distance measures have been proposed for large-scale string processing~\cite{Cormode07,Chakraborty16}.
{\em Edit sensitive parsing~(ESP)}~\cite{Cormode07} is a metric embedding of
a string distance measure called {\em edit distance with moves~(EDM)} that consists of
ordinal edit operations of insertion, deletion and replacement in addition to substring move operation.
ESP maps all the strings from the EDM space into integer vectors named {\em characteristic vectors} in the $L_1$ distance space.
\begin{table*}[t]
\caption{Summary of string alignment kernels.}
\vspace{-0.3cm}
\label{tab:related}
\label{methods}
  \begin{center}
    \setlength{\tabcolsep}{2pt}
  \begin{tabular}{r||c|c|c|c}
                     &         &  {\bf Training} & {\bf Training} & {\bf Prediction}\\
                     & {\bf Approach} & {\bf time} & {\bf space} &  {\bf time} \\
\hline
GAK~\cite{Cuturi07, Cuturi11}  & Global alignment & $O(N^2L^2)$ & $O(N^2)$ & $O(NL^2)$ \\ 
LAK~\cite{Saigo04}   & Local alignment  & $O(N^2L^2)$ & $O(N^2)$ & $O(NL^2)$  \\
D2KE~\cite{Wu18arxiv,Wu18}         & Random feature map & $O(NDL^2)$ & $O(N(L + D))$ &  $O(DL^2)$\\
\hline\hline
SFMEDM (this study) & ESP  & $O(NL + dDN)$  & $O(NL\log{NL} + ND + d)$ & $O(L + dD)$ \\
SFMCGK (this study) & CGK  & $O(NL + dDN)$  & $O(L|\Sigma| + ND + d)$ & $O(L + dD)$ \\

\end{tabular}
\end{center}
\end{table*}
To date, ESP has been applied only to string processing
such as string compression~\cite{Maruyama13}, indexing~\cite{Takabatake14},
edit distance computation~\cite{Cormode07}; however,
as we will see, there remains high potential for application to an approximation of alignment kernels.
ESP is expected to be effective for approximating alignment kernels,
because it approximates EDM between strings as $L_1$ distance between integer vectors.

\smallskip
{\em Contribution}.
In this paper, we present SFMEDM as the first approximation of alignment kernels for 
solving large-scale learning problems on string data. 
Key ideas behind the proposed method are threefold: 
(i) to project input strings into characteristic vectors leveraging ESP,
(ii) to map characteristic vectors into vectors of RFFs by FMs
, and (iii) to train linear SVM on the mapped vectors. 
However, applying FMs for RFFs to high-dimensional vectors in a direct way requires memory linearly proportional to not only dimension $d$ of input vectors but also  dimension $D$ of RFF vectors.
In fact, characteristic vectors as input vectors for FMs tend to be very high dimensional $d$ for solving large-scale problems using FMs, and output vectors of RFFs needs to also be high-dimensional $D$ for achieving high prediction accuracies, and  
those conditions limit the applicability of FMs on a large-scale. 
Although fastfood approach~\cite{Le13} and orthogonal range reporting~\cite{Yu16} have been proposed for efficiently computing RFFs 
in $O(D\log{d})$ time and $O(d)$ memory, they are only applicable to RFFs for approximating RBF kernels with a theoretical guarantee. 
Accordingly, in this study, we present {\em space-efficient FMs (SFMs)} that requires only $O(d)$ memory to solve this problem and can be used for approximating any shift-invariant kernel such as a Laplacian kernel. This is an essential property which is required for approximating alignment kernels and has not been taken into account by previous research.
Our SFMEDM has the following desirable properties: 
\begin{enumerate}
 \setlength{\parskip}{0cm}
 \setlength{\itemsep}{0cm}
\item {\bf Scalability: } SFMEDM is applicable to massive string data.
\item {\bf Fast training: } SFMEDM trains SVM fast.
\item {\bf Space efficiency: } SFMEDM trains SVM space-efficiently.
\item {\bf Prediction accuracy: } SFMEDM can achieve high prediction accuracy.
\end{enumerate}
We experimentally test the ability of SFMEDM to train SVM with various massive string data, and 
demonstrate that SFMEDM has superior performance in terms of prediction accuracy, scalability and computational efficiency. 



\section{Related Work}

Several alignment kernels have been proposed for analyzing string data. 
We briefly review the state of the art, which is also summarized in Table~\ref{tab:related}.
Early methods are proposed in \cite{Bahlmann02, Shimodaira02, Zhou10} and 
are known not to satisfy the positive definiteness for their kernel matrices. 
Thus, they are proposed with numerical corrections for any deficiency of the kernel matrices. 

The {\em global alignment kernel~(GAK)}~\cite{Cuturi07,Cuturi11} is an alignment kernel based on global alignments originally proposed for time series data.
GAK defines a kernel as a summation score of all possible global alignments between two strings. 
The computation time of GAK is $O(N^2L^2)$  for number of strings $N$ and 
the length of strings $L$, and its space usage is $O(N^2)$.

A {\em local alignment kernel~(LAK)} on the notion of the Smith-Waterman algorithm~\cite{Smith81} for detecting protein remote homology 
was proposed by Saigo et al.~\cite{Saigo04}.
LAK measures the similarity between each pair of strings by summing up scores obtained 
from local alignments with gaps of strings. 
The computation time of LAK is $O(N^2L^2)$ and its space usage is $O(N^2)$. 
Although, in combination with SVM, LAK achieves high classification accuracies for protein sequences, LAK is applicable to protein strings only 
because its scoring function is optimized for proteins. 

{\em D2KE}~\cite{Wu18arxiv} is a random feature map from structured data to feature vectors such that a distance measure between each pair of 
the structured data is preserved by the inner product between the corresponding pair of mapped vectors. 
The feature vector for each input structured data is built as follows: 
(i) $D$ structured data in input are sampled; (ii) the $D$-dimension feature vector for each structured data is built such that each dimension of the feature vector is defined as the distance between the structured data and a sampled one. 
D2KE has been applied to time series data~\cite{Wu18}; however, as we will see, D2KE cannot achieve high prediction accuracies when it is applied to string data.

Despite the importance of a scalable learning with alignment kernels, no previous work has been able to achieve high scalabilities while preserving high prediction accuracies.
We present SFMEDM, the first scalable learning with string alignment kernels that meets these demands and is made possible by leveraging an idea behind ESP and SFM.

{\em CGK}~\cite{Chakraborty16} is another metric embedding for edit distance and maps input strings $S_i$ of alphabet $\Sigma$ and of the maximum length $L$
into strings $S^\prime_i$ of fixed-length $L$ such that the edit distance between each pair of input strings is approximately preserved by
the Hamming distance between the corresponding pair of mapped strings.
Recently, CGK has been applied to the problem of edit similarity joins~\cite{Zhang17}.
We also present a kernel approximation of alignment kernels called {\em SFMCGK} by leveraging an idea behind CGK and SFM.

Details of the proposed method are presented in the next section.

\section{Edit sensitive parsing}
{\em Edit sensitive parsing~(ESP)}~\cite{Cormode07} is an approximation method for efficiently computing 
{\em edit distance with moves~(EDM)}. 
EDM is a string-to-string distance measure for turning one string into another in a series of string operations, where 
a substring move is included as a string operation in addition to typical string operations such as insertion, deletion and replacement.  
Formally, let $S$ be a string of length $L$ and $S[i]$ be the $i$-th character in $S$. 
$EDM(S,S^\prime)$ for two strings $S$ and $S^\prime$ is defined as the minimum number of edit operations defined below to transform $S$ into $S^\prime$ as following: \\
{\bf Insertion:} character $a$ at position $i$ in $S$ is inserted, resulting in $S[1]...S[i-1]aS[i]S[i+1]...S[L]$; \\
{\bf Deletion:}  character $S[i]$ at position $i$ in $S$ is deleted, resulting in $S[1]...S[i-1]S[i+1]...S[L]$; \\
{\bf Replacement:} character $S[i]$ at position $i$ in $S$ is replaced by $a$, resulting in $S[1]...S[i-1]aS[i+1]...S[L]$; \\
{\bf Substring move:} a substring $S[i]S[i+1]...S[j]$ in $S$ is moved and inserting at position $p$, resulted in $S[1]...S[i-1]S[j+1]...S[p-1]S[i]...S[j]S[p]...S[L]$. \\
%
Computing EDM between two strings is known as an NP-complete problem~\cite{Shapira02}.
ESP can approximately compute EDM by embedding strings into $L_1$ vector space by a parsing.

\begin{figure}
\begin{center}
\includegraphics[width=0.4\textwidth]{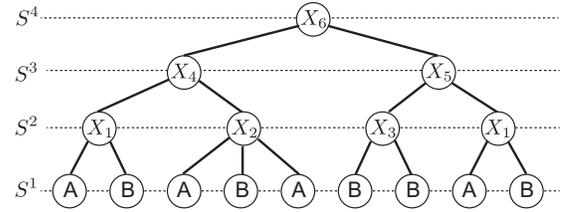}
\end{center}
\vspace{-0.4cm}
\caption{Illustration of an ESP tree for string $S=ABABABBAB$.}
\label{fig:exptree}
\end{figure}

Given string $S$, ESP builds a parse tree named an {\em ESP tree}, 
which is illustrated in Figure~\ref{fig:exptree} as an example. 
The ESP tree is a balanced tree and each node in the ESP tree belongs to one of three types: 
(i) a node with three children, (ii) a node with two children and (iii) a node without children (i.e., a leaf).
In addition, internal nodes in the ESP tree have the same node label if and only if they have children satisfying both two conditions: 
(i) the numbers of those children are the same, and (ii) the node labels of those children are the same in the left-to-right order. 
The height of ESP tree is $O(\log{L})$ for the length of input string $L$. 

Let $V(S)\in \N^d$ be a $d$-dimension integer vector built from ESP tree $T(S)$ such that each dimension of $V(S)$ is the number of a node label appearing in $T$. 
$V(S)$ is called {\em characteristic vectors}.
ESP builds ESP trees such that as many subtrees with the same node labels as possible are built for common substrings for strings $S$ and $S^\prime$, 
resulted in an approximation of EDM between $S$ and $S^\prime$ by $L_1$ distance between their characteristic vectors $V(S)$ and $V(S^\prime)$, 
i.e., $EDM(S, S^\prime) \approx ||V(S) - V(S^\prime)||_1$, where $||\cdot||_1$ is an $L_1$ norm.
More precisely, the upper  and lower bounds of the approximation are as follows, 
\begin{eqnarray}
  EDM(S,S^\prime) \leq ||V(S) - V(S^\prime)||_1 \nonumber \\
  \leq O(\log{L}\log^*{L})EDM(S,S^\prime),  \nonumber
\end{eqnarray}
where $\log^*{L}$ is the iterated logarithm of $L$, which is recursively defined as 
$\log^{1}L=\log_2{L}$, $\log^{i+1}{L}=\log\log^i{L}$ and $\log^*{L}= \min\{k; \log^k{L} \leq 1, i \geq 1\}$ for a positive integer $L$. 

Detail of the ESP algorithm is presented in the appendix. 

\section{Space-efficient Feature Maps}\label{sec:linear}
In this section we present our new SFMs for RFFs using space proportional to the dimension $d$ of characteristic vectors and 
\emph{independent} of the RFF target dimension $D$. 
The proposed SFMs improve space usage for generating RFFs from $O(dD)$ to $O(d)$ while preserving theoretical guarantees (concentration bounds).
The method is general and can be used for approximating any shift-invariant kernel.

From an abstract point of view, an RFF is based on a way of constructing a random mapping
\[
z_{\mathbf r}: \R^d \rightarrow [-1,+1]^2
\]
such that for every choice of vectors $\x,\y\in\R^d$ we have 
\[
\E[z_{\mathbf r}(\x)^\prime z_{\mathbf r}(\y)] = k(\x,\y),
\]
where $k$ is the kernel function.
The randomness of $z_{\mathbf r}$ comes from a vector ${\mathbf r}\in\R^d$ sampled from an appropriate distribution $\D_k$ that depends on kernel function~$k$ (see section~\ref{sec:kernel} for more details), and the expectation is over the choice of ${\mathbf r}$.
For the purposes of this section, all that needs to be known about $\D_k$ is that the $d$ vector coordinates are independently sampled according to the marginal distribution~$\Delta_k$.

Since $(z_{\mathbf r}(\x)^\prime z_{\mathbf r}(\y))^2 \leq 1$ we have $Var(z_{\mathbf r}(\x)^\prime z_{\mathbf r}(\y))\leq 1$, i.e., bounded variance; however, this in itself does not imply the desired approximation as $k(\x,\y)\approx z_{\mathbf r}(\x)^\prime z_{\mathbf r}(\y)$. Indeed, $z_{\mathbf r}(\x)^\prime z_{\mathbf r}(\y)$ is a poor estimator of $k(\x,\y)$.
The accuracy of RFFs can be improved by increasing the output dimension to $D \geq 2$.
Specifically RFFs use $D/2$ \emph{independent} vectors ${\mathbf r}_1,\dots,{\mathbf r}_{D/2}\in\R^d$ sampled from $\D_k$, and they consider FMs 
\[
 \z: \x \mapsto \sqrt{\tfrac{2}{D}} \left(z_{{\mathbf r}_1}(\x),z_{{\mathbf r}_2}(\x),\dots,z_{{\mathbf r}_{D/2}}(\x)\right)
\]
that concatenates the values of $D/2$ functions to one $D$-dimensional vector.
It can then be shown that $|\z(\x)^\prime \z(\y) - k(\x,\y)| \leq \varepsilon$ with high probability for sufficiently large $D = \Omega(1/\varepsilon^2)$.

To represent the function $\z$, it is necessary to store a matrix containing vectors ${\mathbf r}_1,\dots,{\mathbf r}_{D/2}$, which uses space $O(dD)$.
Our assumption for ensuring good kernel approximations is that the vectors ${\mathbf r}_i$ do not need to be independent.
Instead, for a small integer parameter $t\in N$, we {\emph compute} each vector ${\mathbf r}_i$ using a hash function $h: \{1,\dots,D/2\} \rightarrow \R^d$ chosen from a $t$-wise independent family such that for every $i$, $h(i)$ comes from distribution~$\D_k$.
Then, instead of storing ${\mathbf r}_1,\dots,{\mathbf r}_{D/2}$, we only store the description of the hash function $h$ in memory $O(td)$.
A priori, two issues seemingly concern this approach:
 \vspace{-0.5\baselineskip}           
{
\setlength{\leftmargini}{0pt}         
\begin{itemize}
  \setlength{\itemsep}{2pt}      
  \setlength{\parskip}{0pt}      
  \setlength{\itemindent}{30pt}   
  \setlength{\labelsep}{5pt}     
\item It is unclear how to construct $t$-wise independent hash functions with output distribution $\D_k$.
\item Is $t$-wise independence sufficient to ensure results similar to the fully independent setting?
\end{itemize}
}
\vspace{-0.5\baselineskip}
We address these issues in the next two subsections.

\subsection{Hash functions with distribution $\D_k$}
For concreteness, our construction is based on the following class of $t$-wise independent hash functions, where $t\in\N$ is a parameter:
For $\av = (a_0,a_1,\dots,a_{t-1})\in [0,1]^t$ chosen uniformly at random, let
\[ f_{\av}(x) = \sum_{j=0}^{t-1} a_j x^j \text{ mod } 1 \]
where $y \text{ mod } 1$ computes the fractional part of $y \in \R$.
It can be shown that any $t$ distinct integer inputs $i_1,\dots,i_t \in \N$, the vector $(f_{\av}(i_1),\dots,f_{\av}(i_t))$ is uniformly distributed in $[0,1]^t$.

Let $CDF^{-1}$ denote the inverse of the cumulative distribution function of the marginal distribution~$\Delta_k$.
Then, if $y$ is uniformly distributed in $[0,1]$, $CDF^{-1}(y)\sim \Delta_k$.
Accordingly hash function $h$ can be constructed where the $j$-th coordinate on input $i$ is given, as
\[ h(i)_j = CDF^{-1}(f_{\av^j}(i)) \]
where $\av^1,\dots,\av^d$ are chosen independently from $[0,1]^d$. 
We see that for every $i\in\N$, $h(i) = (h(i)_1,\dots,h(i)_d)$ has distribution $\D_k$.
Furthermore, for every set of $t$ distinct integer inputs $i_1,\dots,i_t \in \N$, the hash values $h(i_1),\dots,h(i_t)$ are independent.

\subsection{Concentration bounds}
We then show that for RFFs, $D=O(1/\varepsilon^2)$ random features suffice to approximate the kernel function within error $\varepsilon$ 
with probability arbitrarily close to $1$.
\begin{thm}\label{thm:main}
For every pair of vectors $\x,\y\in\R^d$, if the mapping $\z$ is constructed as described above using $t\geq 2$, for every $\varepsilon > 0$, it follows that 
\[ \Pr[|\z(\x)^\prime \z(\y) - k(\x,\y)| \geq \varepsilon] \leq 2/(\varepsilon^2 D)  \enspace .\]
\end{thm}

\paragraph*{Proof}
       Our proof follows the same outline as the standard proof of Chebychev's inequality.
        Consider the second central moment:
        \begin{align*}
        & \E\left[\left(z(\x)^\prime z(\y) - k(\x,\y)\right)^2\right] \\
        & = \E\left[\left(\tfrac{2}{D} \sum_{i=1}^{D/2} z_{h(i)}(\x)^\prime z_{h(i)}(\y) - k(\x,\y)\right)^2\right] \\
        & = \E\left[\sum_{i=1}^{D/2} \left( z_{h(i)}(\x)^\prime z_{h(i)}(\y) - \tfrac{D}{2}\, k(\x,\y)\right)^2\right] \\
        & = \tfrac{4}{D^2} \sum_{i=1}^{D/2} \E\left[ \left( z_{h(i)}(\x)^\prime z_{h(i)}(\y) - \tfrac{D}{2}\, k(\x,\y)\right)^2 \right] \leq 2/D \enspace .
        \end{align*}

The second equality above uses $2$-wise independence, and the fact that
\[
\E\left[\sum_{i=1}^{D/2} z_{h(i)}(\x)\cdot z_{h(i)}(\y) - \tfrac{D}{2} k(\x,\y)\right] = 0
\]
 to conclude that only $D/2$ terms in the expansion have nonzero expectation.
Finally, we have:

\begin{align*}
        & \Pr[|\z(\x)^\prime \z(\y) - k(\x,\y)| \geq \varepsilon] \\
        & \leq \Pr[(\z(\x)^\prime \z(\y) - k(\x,\y))^2 \geq \varepsilon^2]\\
        & \leq \E[(\z(\x)^\prime \z(\y) - k(\x,\y))^2]/\varepsilon^2 \leq 2/(\varepsilon^2 D),
\end{align*}

where the second inequality follows from Markov's inequality. This concludes the proof.

%
%
%

In the original analysis of RFFs, a strong approximation guarantee was considered; namely, the kernel function for \emph{all} pairs of points $\x,\y$ in a bounded region of $\R^d$ was approximated.
This kind of result can be achieved by choosing $t\geq 2$ sufficiently large to obtain strong tail bounds.
However, we show that the point-wise guarantee (with $t=2$) provided by Theorem~\ref{thm:main} is sufficient for an application in kernel approximations in Sec.~\ref{sec:exp}.

\begin{algorithm}[t]
\caption{Generation of Cauchy random numbers using $2$-wise independent hash function. $array_1$, $array_2$: arrays of $d$ 64-bit unsigned integers; $UMAX32$: maximum value of unsigned 32-bit integer; $\beta$: a parameter.}
\label{alg:hashing}
\begin{algorithmic}[1]
\State Initialize $array_1$ and $array_2$ with 64-bit random numbers as unsigned integers.
\Function{Func\_f}{$i,j$}
\State $f = array_1[j] + array_2[j]\cdot i$ \Comment{Compute hash value}
\State $v = f >> 32$ \Comment{Get the most-significant 32-bit of value}
\State \Return{$v/UMAX32$} \Comment{Return value in $[0,1]$}
\EndFunction
\Function{Func\_h}{$i,j$}
\State $u = Func\_F(i,j)$
\State \Return{$tan(\pi\cdot (u - 0.5))/\beta$} \Comment{Convert random number $u$ to Cauchy random number}
\EndFunction
\end{algorithmic}
\end{algorithm}
\begin{algorithm}[t]
\caption{Construction of RFFs by SFMs. $\z$: vector of RFFs; $D$: dimension of $\z$; $V$: characteristic vector; $d$: dimension of $V$.}
\label{alg:RFF}
\begin{algorithmic}[1]
\Function{SFM}{$V$}
\For{$i=1,...,D/2$}
\State $s = 0$
\For{$j=1,...,d$}
\State $s=s+V[j]\cdot Func\_H(i,j)$
\EndFor
\State $\z[2\cdot i - 1] = \sqrt{\frac{2}{D}}\cdot sin(s)$
\State $\z[2\cdot i] = \sqrt{\frac{2}{D}}\cdot cos(s)$
\EndFor
\State \Return{$\z$}
\EndFunction
\end{algorithmic}
\end{algorithm}

\begin{table*}[t]
\caption{Summary of datasets.}
\label{tab:dataset}
\vspace{-0.2cm}
\begin{center}
\begin{tabular}{l|rrrr}
{\bf Dataset} & {\bf Number} & {\bf \#positives} & {\bf Alphabet size} & {\bf Average length}  \\
\hline 
Protein & 3,238 & 96 &20 & 607  \\
DNA     & 3,238 & 96 & 4 & 1,827 \\
Music & 10,261 & 9,022 & 61 & 329 \\
Sports & 296,337 & 253,017 & 63 & 307 \\
Compound & 1,367,074 & 57,536 & 44 &  53  \\
\end{tabular}
\end{center}
\end{table*}
\begin{table*}
\begin{center}
\caption{Execution time in seconds, memory in mega bytes and dimension $d$ of characteristic vectors by ESP and CGK for each dataset.}
\vspace{-0.3cm}
\begin{tabular}{|l|c|c|c|c|c|c|c|c|c|c|}
\hline
Data & \multicolumn{2}{c|}{\bf Protein} & \multicolumn{2}{c|}{\bf DNA} & \multicolumn{2}{c|}{\bf Music} & \multicolumn{2}{c|}{\bf Sports} & \multicolumn{2}{c|}{\bf Compound}\\
\hline
Method     & ESP & CGK & ESP & CGK & ESP & CGK & ESP & CGK & ESP & CGK \\
\hline
\hline
Time~(sec) & $1.25$ & $0.87$ & $2.01$ & $2.63$ & $1.86$ & $0.86$ & $47.83$ & $34.08$ & $32.73$ & $28.70$ \\
Memory~(MB) & $1,042.90$ & $0.09$ & $1,049.89$ & $3.38$ & $1,048.30$ & $0.37$ & $1,514.23$ & $0.54$ & $1,165.60$ & $0.08$\\
Dimension $d$ & $707,922$ & $4,950,686$ & $969,653$ & $19,192,656$ & $910,110$ & $2,129,505$ & $18,379,173$ &$3,095,844$ & $5,302,660$ & $485,840$ \\
\hline
\end{tabular}
\label{tab:espcgk}
\end{center}
\end{table*}

\section{Scalable Alignment Kernels}\label{sec:kernel}
We present the SFMEDM algorithm for scalable learning with alignment kernels hereafter. 
Let us assume a collection of $N$ strings and their labels $(S_1,y_1),$ $(S_2,y_2), ..., (S_N,y_N)$ where $y_i\in \{0,1\}$. 
We define alignment kernels using $EDM(S_i, S_j)$ for each pair of strings $S_i$ and $S_j$ as follows, 
\begin{eqnarray}
k(S_i,S_j)=\exp(-EDM(S_i,S_j)/\beta), \nonumber
\end{eqnarray}
where $\beta$ is a parameter. 
We apply ESP to each $S_i$ for $i=1,2,...,N$ and build ESP trees $T_1, T_2,...,T_N$.
Since ESP approximates $EDM(S_i,S_j)$ as an $L_1$ distance between characteristic vectors $V(S_i)$ and $V(S_j)$ built from ESP trees 
$T_i$ and $T_j$ for $S_i$ and $S_j$, i.e., $EDM(S_i,S_j)\approx ||V(S_i)-V(S_j)||_1$, 
$k(S_i,S_j)$ can be approximated as follows, 
\begin{eqnarray}\label{eq:kernel}
k(S_i,S_j) \approx \exp(-||V(S_i)-V(S_j)||_1/\beta). 
\end{eqnarray}
Since Eq.\ref{eq:kernel} is a Laplacian kernel, which is also known as a shift-invariant kernel~\cite{Rahimi07}, 
we can approximate $k(S_i,S_j)$ using FMs $\z(\x)$ for RFFs as follows,
\begin{eqnarray}
k(S_i,S_j)\approx \z(V(S_i))^\prime \z(V(S_j)),  \nonumber
\end{eqnarray}
where 
$\z(\x) = \sqrt{\frac{2}{D}}(\z_{\mathbf r_1}(\x),\z_{\mathbf r_2}(\x),...,\z_{\mathbf r_{D/2}}(\x))$. 
For Laplacian kernels, $\z_{{\mathbf r_m}}(\x)$ for each $m=1,2,...,D/2$ is defined as
\begin{eqnarray}\label{eq:rff}
  \z_{{\mathbf r_m}}(\x) = (\cos{({\mathbf r}^\top_m\x)}, \sin{({\mathbf r}^\top_m\x)}) 
\end{eqnarray}
where random vectors ${\mathbf r}_m \in \R^d$ for $m=1,2,...,D/2$ are sampled from the Cauchy distribution. 
We shall refer to approximations of alignment kernels leveraging ESP and FMs as {\em FMEDM}. 

Applying FMs to high dimensional characteristic vectors consumes $O(dD)$ memory for storing vectors ${\mathbf r}_m \in \R^d$ for $m=1,2,...,D/2$. 
Thus, we present SFMs for RFFs using only $O(td)$ memory by applying $t$-wise independent hash functions introduced in Sec.~\ref{sec:linear}.
We fix $t=2$ in this study, resulted in $O(d)$ memory. 
We shall refer to approximations of alignment kernels leveraging ESP and SFMs as {\em SFMEDM}.

Algorithm~\ref{alg:hashing} generates random numbers from a Cauchy distribution by using $O(d)$ memory. 
Two arrays $array_1$ and $array_2$, initialized with 64-bit random numbers as unsigned integers, are used. 
Function $f_a(x)$ is implemented using $array_1$ and $array_2$ in $Func\_F$ and returns a random number in $[0, 1]$ for given $i$ and 
$j$ as input. 
Then, random number $u$ returned from $Func\_F$ is converted to a random number from the Cauchy distribution in $Func\_H$ as 
$\tan(\pi \cdot (u - 0.5))/\beta$ at line~8. 
Algorithm~\ref{alg:RFF} implements SFMs generating RFFs in Eq.\ref{eq:rff}. 
Computation time and memory for SFMs are $O(dDN)$ and $O(d)$, respectively. 



\begin{figure*}[t]
\begin{center}
\begin{tabular}{ccccc}
{\small \bf Protein} & {\small\bf DNA} & {\small\bf Music} & {\small\bf Sports} & {\small\bf Compound}\\

\includegraphics[width=0.17\textwidth]{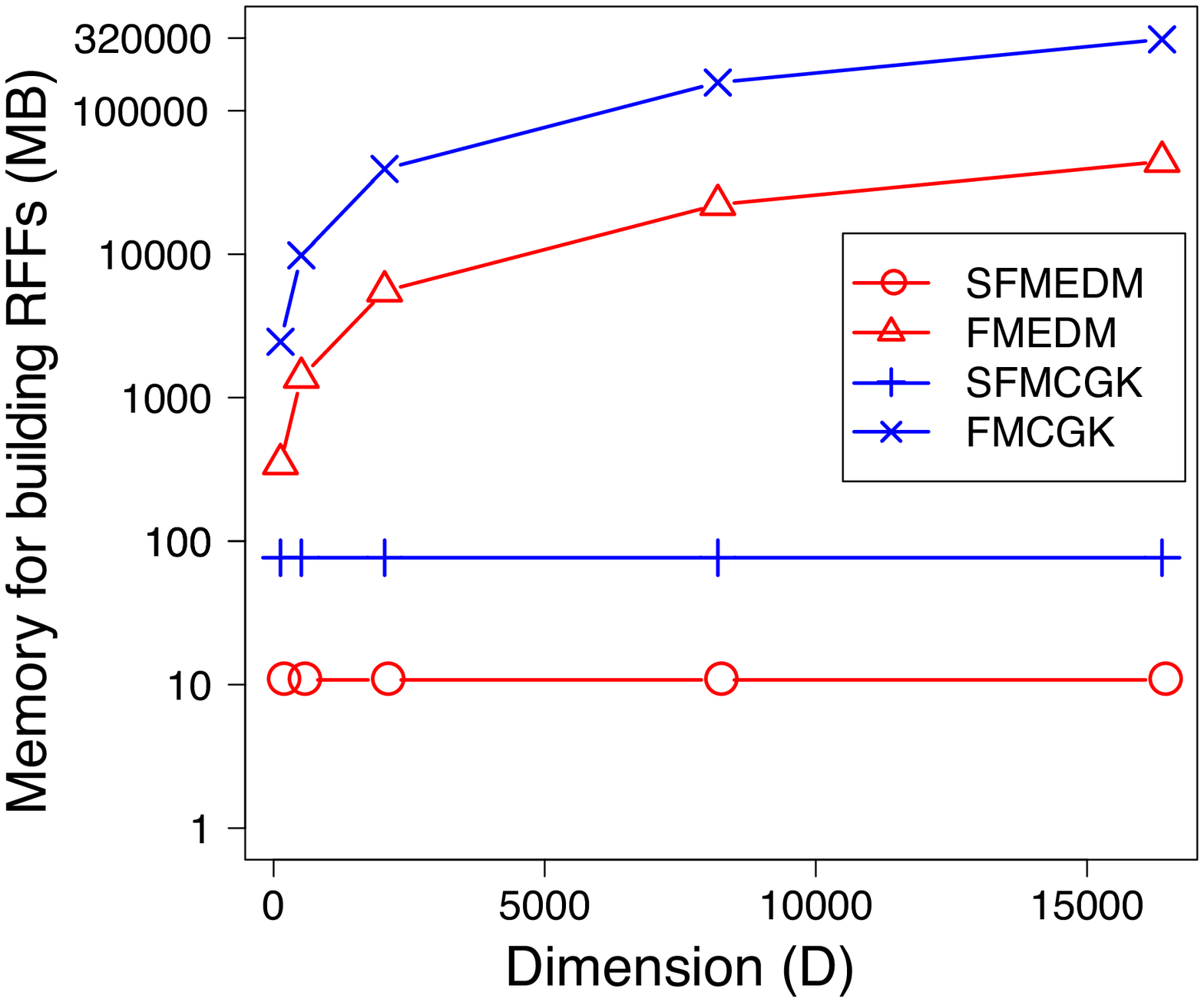} &
\includegraphics[width=0.17\textwidth]{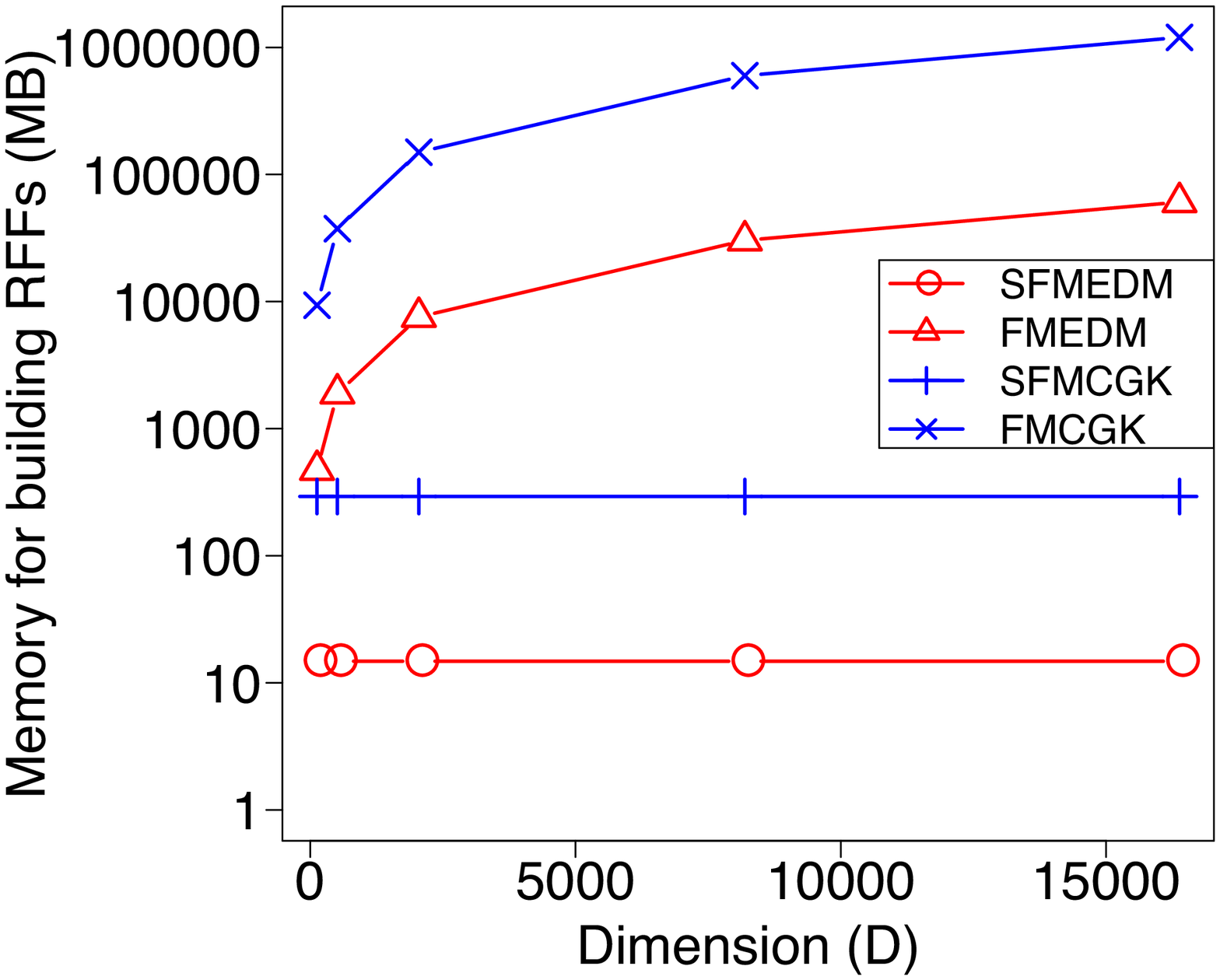} &
\includegraphics[width=0.17\textwidth]{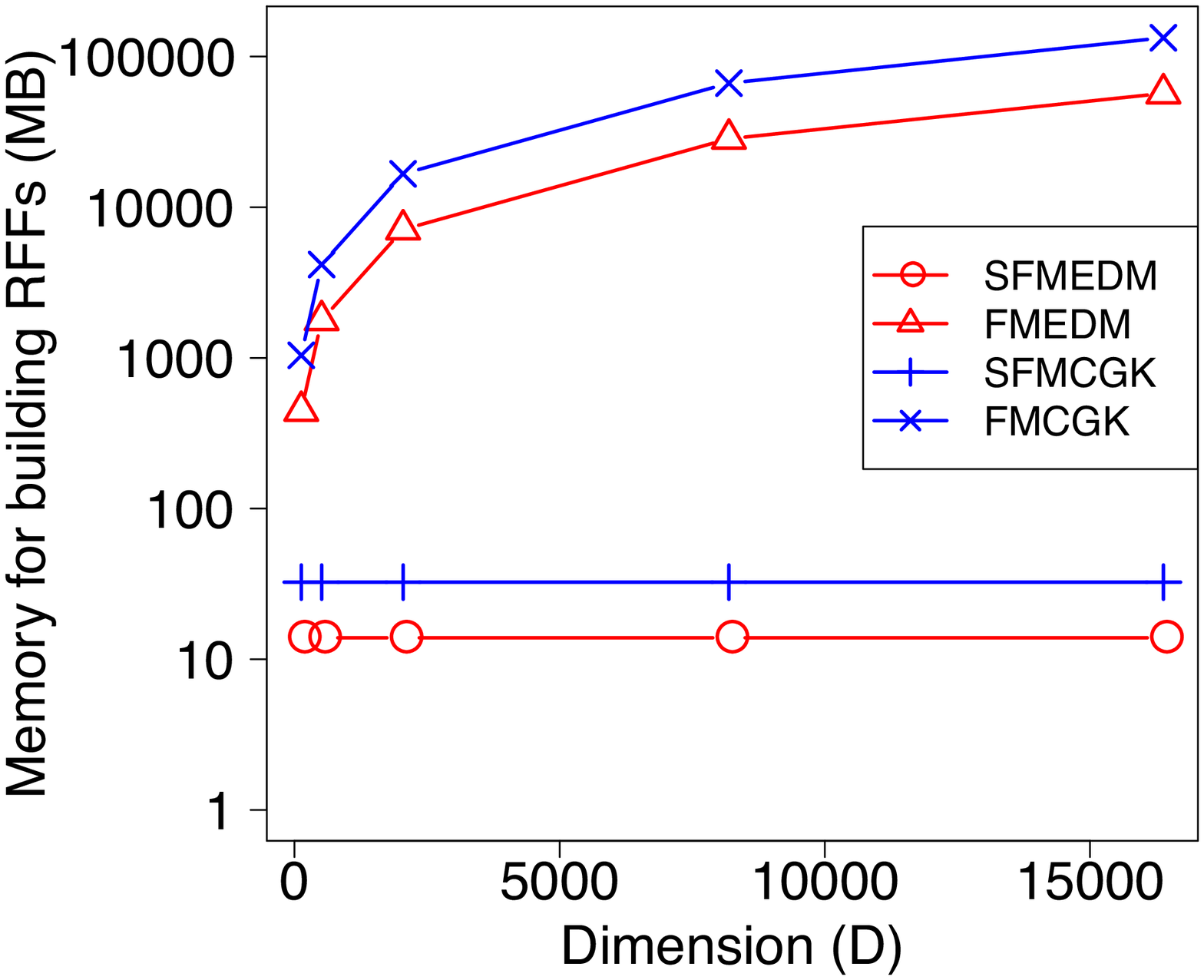} &
\includegraphics[width=0.17\textwidth]{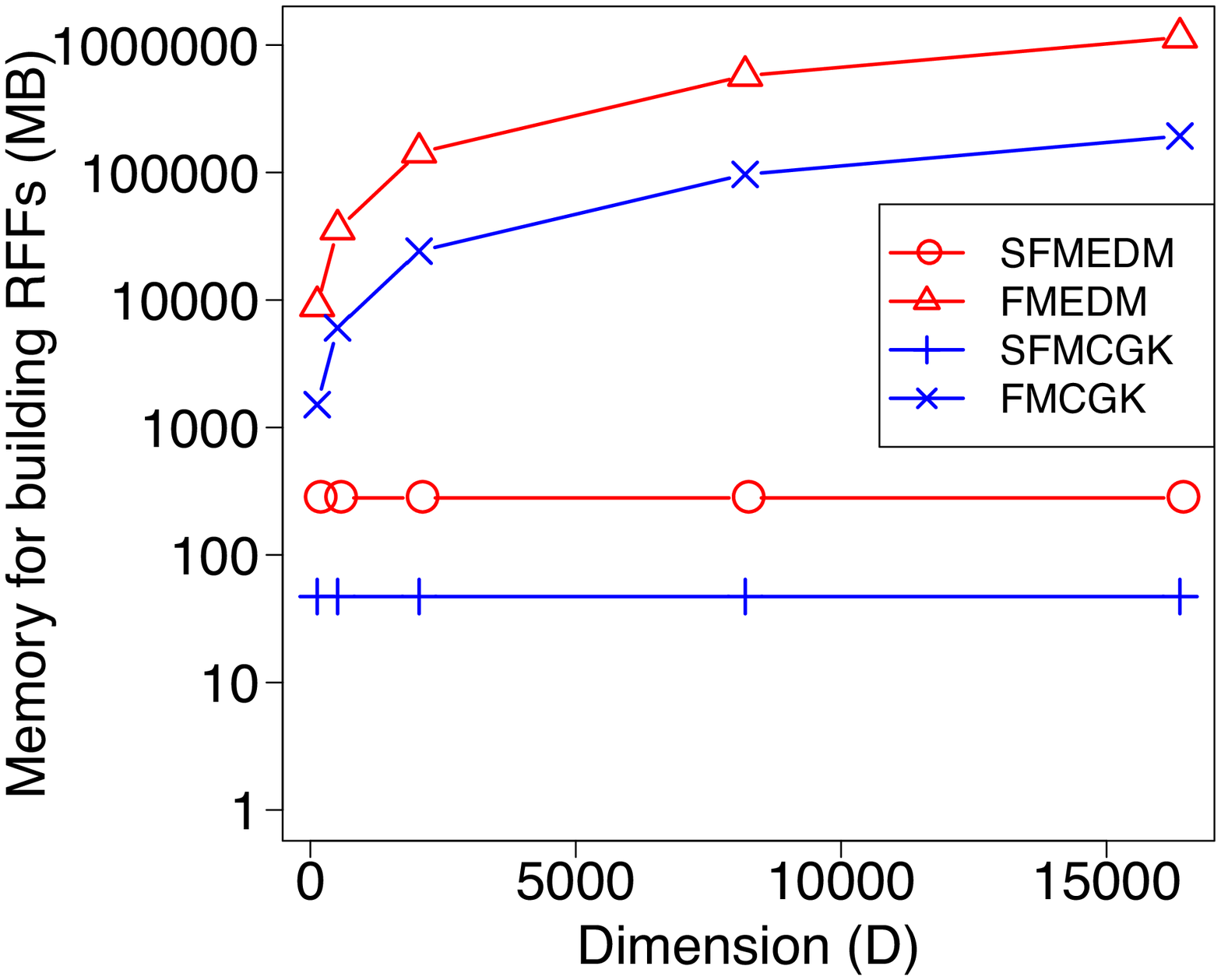} &
\includegraphics[width=0.17\textwidth]{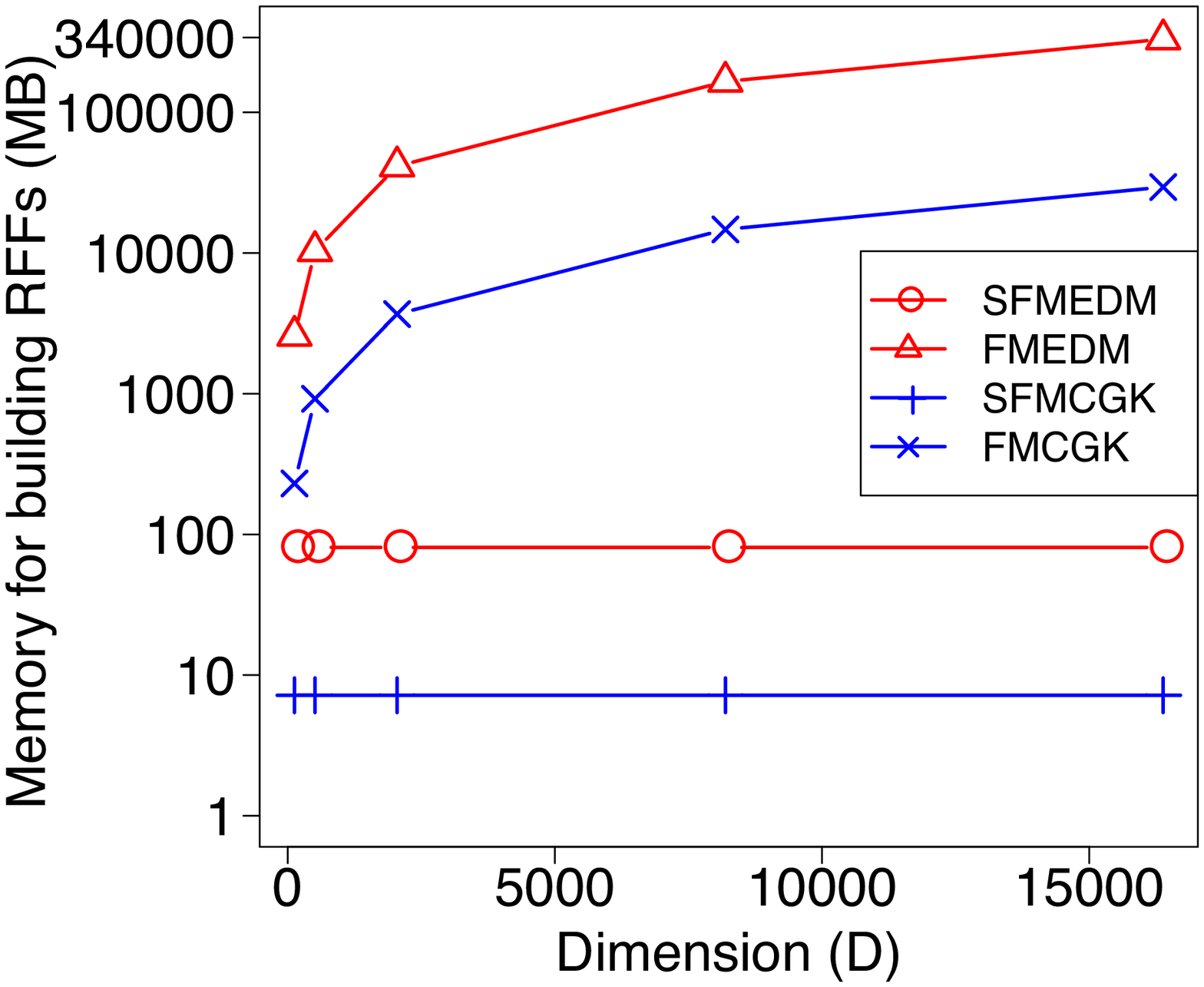} 
\end{tabular}
\end{center}
\vspace{-0.5cm}
\caption{Memory in megabytes for building vectors of RFFs for various dimensions $D$.}
\label{fig:rffmemory}
\begin{center}
\begin{tabular}{ccccc}
{\small\bf Protein} & {\small\bf DNA} & {\small\bf Music} & {\small\bf Sports} & {\small\bf Compound}\\
\includegraphics[width=0.17\textwidth]{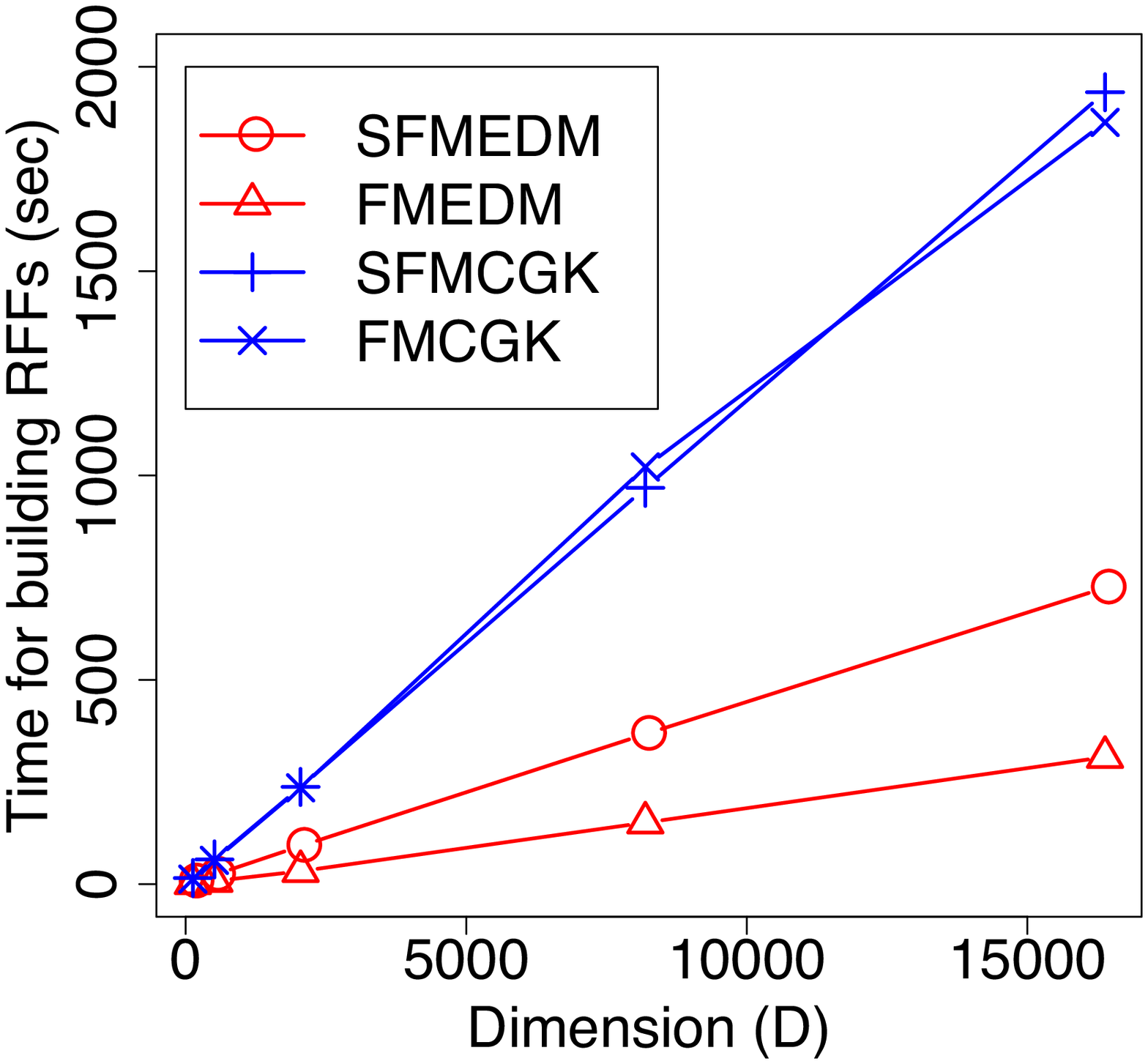} &
\includegraphics[width=0.17\textwidth]{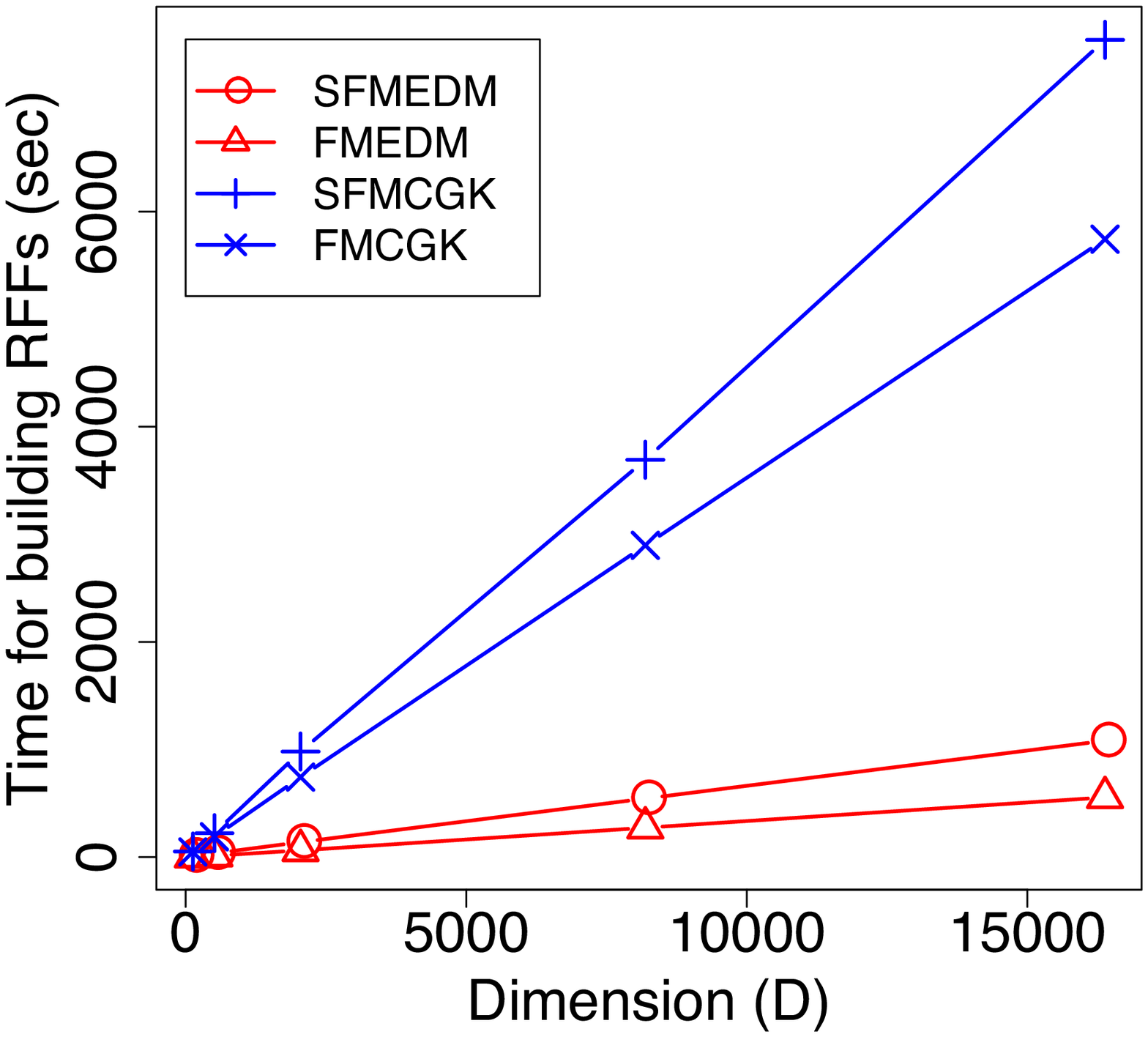} &
\includegraphics[width=0.17\textwidth]{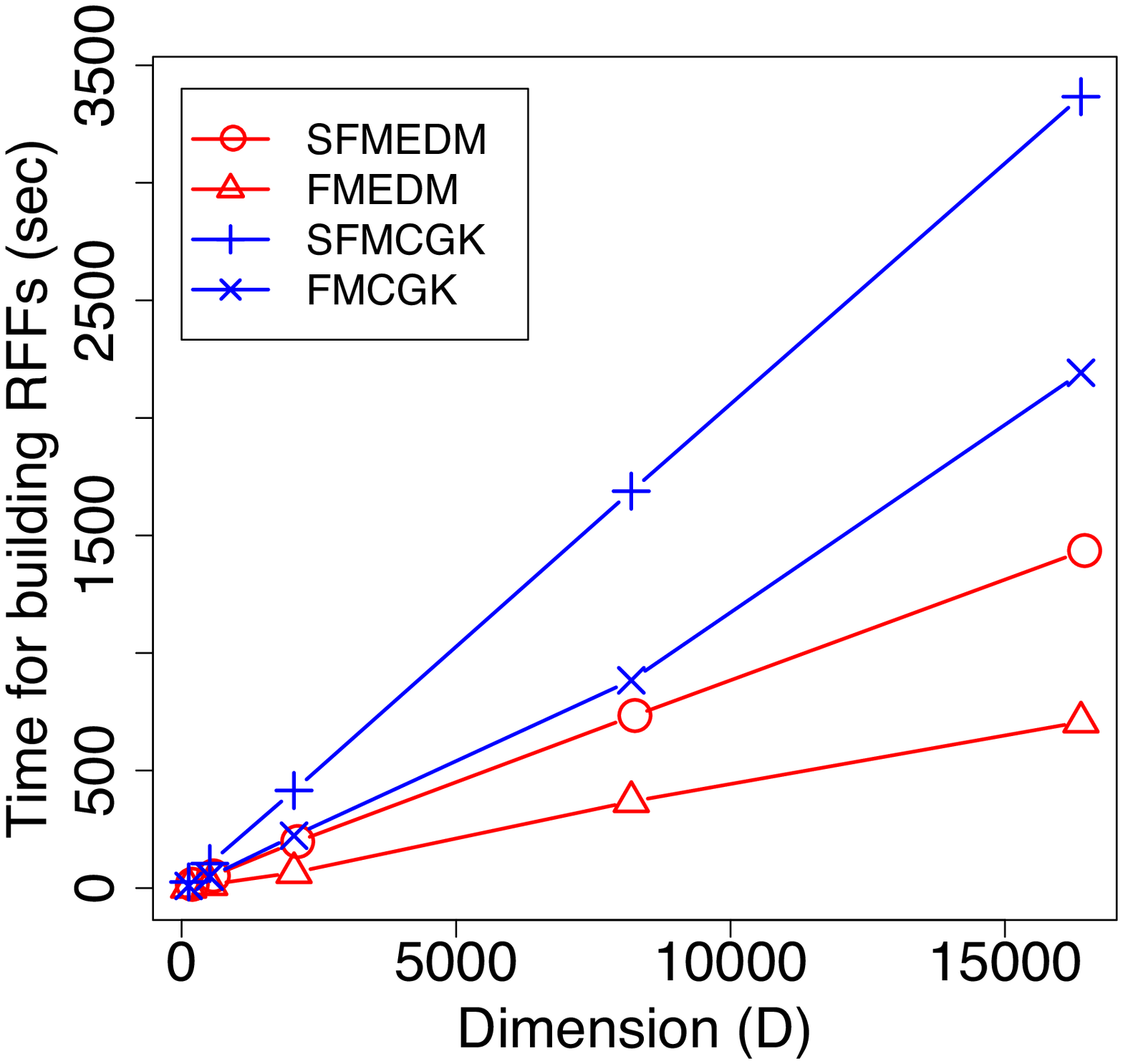} &
\includegraphics[width=0.17\textwidth]{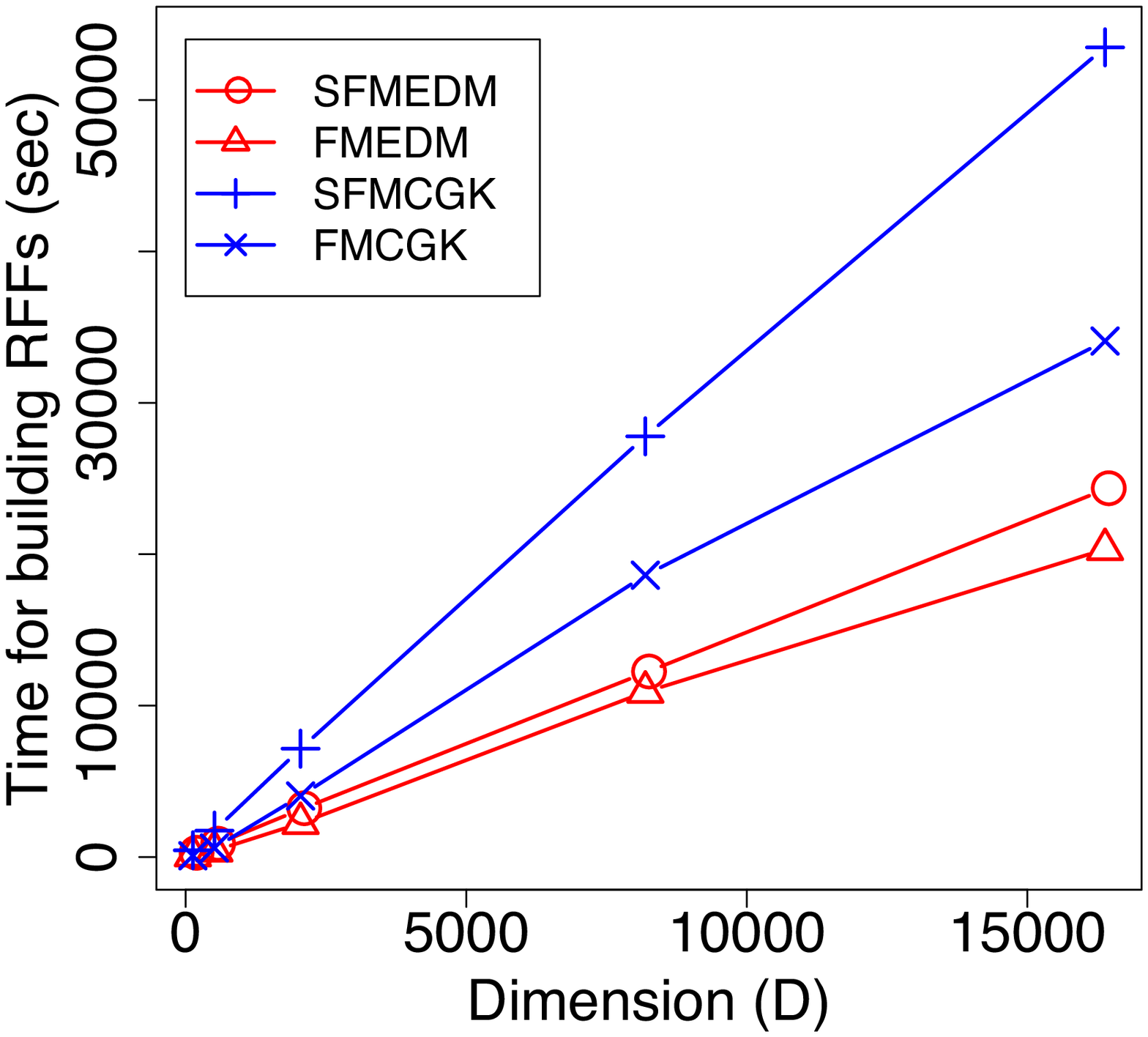}  &
\includegraphics[width=0.17\textwidth]{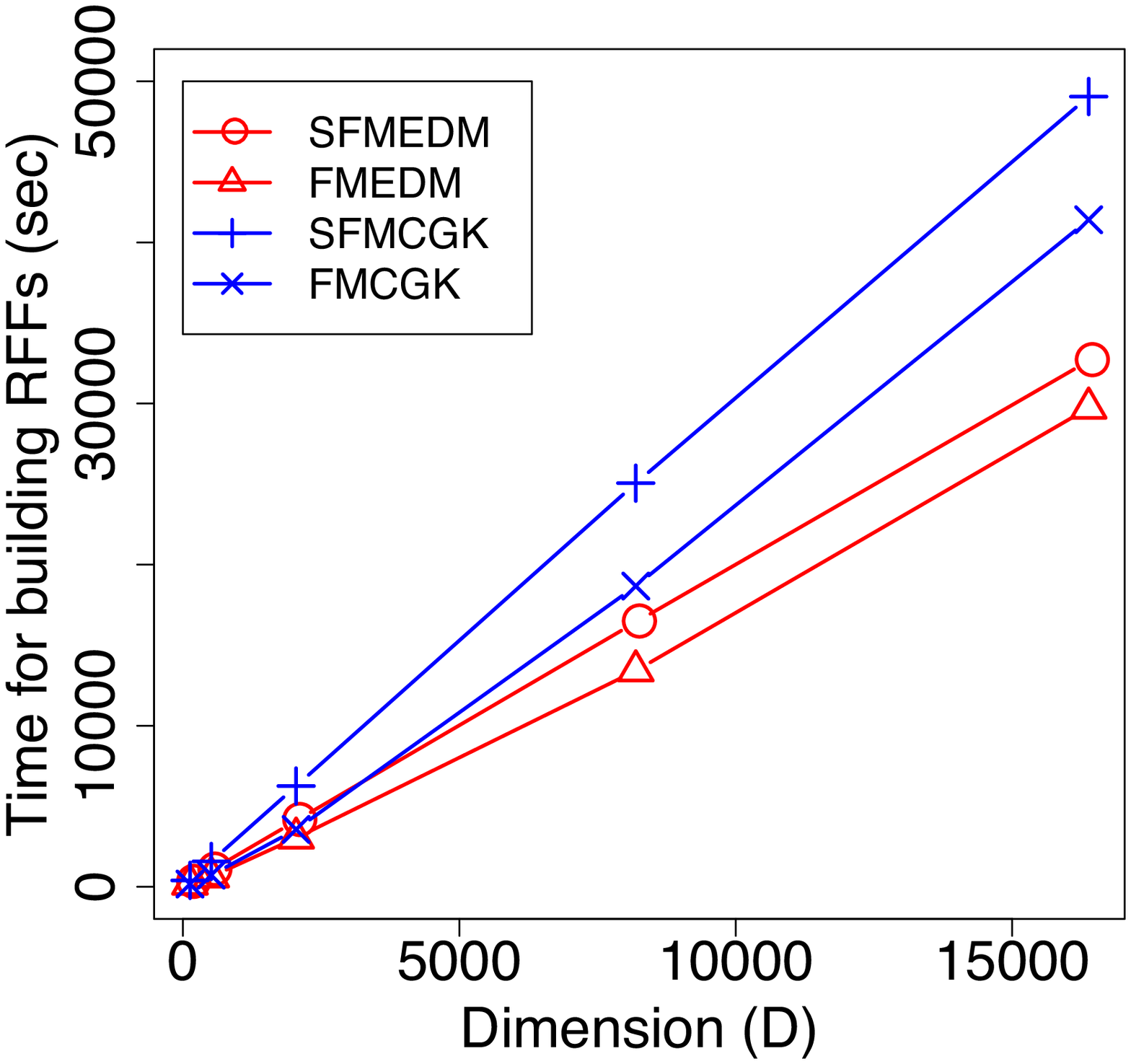} 
\end{tabular}
\end{center}
\vspace{-0.5cm}
\caption{Time in seconds for building vectors of RFFs for various dimensions $D$.}
\label{fig:rfftime}
\end{figure*}
\begin{table*}[t]
\begin{center}
\caption{Average error by SFMEDM and FMEDM for each dataset. All values are multiplied by $10^2$.}
\begin{tabular}{|l||c|c|c|c|c|}
\hline
{\bf Method}  & {\bf Protein} & {\bf DNA} & {\bf Music} & {\bf Sports} & {\bf Compound}\\
\hline\hline
SFMEDM(D=128) & $7.054(\pm 5.320)$ & $7.051(\pm 5.318)$  & $7.058(\pm 5.318)$ & $7.058(\pm 5.319)$ & $7.057(\pm 5.317)$ \\
FMEDM(D=128)  & $7.055(\pm 5.321) $ & $7.054 (\pm 2.664)$ &  $7.059(\pm 5.318)$& $7.059(\pm 5.320)$ & $7.057(\pm 5.318)$\\
\hline
SFMEDM(D=512) & $3.523(\pm 2.662) $ & $3.526(\pm 2.663)$ & $3.526(\pm 2.662)$ & $3.525(\pm 2.662)$ & $3.527(\pm 2.662)$\\
FMEDM(D=512)  & $3.526(\pm 2.665)$ & $3.526(\pm 2.666)$  & $3.526(\pm 2.663)$ & $3.526(\pm 2.663)$ & $3.526(\pm 2.662)$\\
\hline
SFMEDM(D=2048) & $1.762(\pm 1.332)$ &  $1.762(\pm 1.332)$ & $1.762(\pm 1.332)$ & $1.762(\pm 1.332)$ & $1.761(\pm 1.331)$\\
FMEDM(D=2048)  & $1.763(\pm 1.332)$ &  $1.762(\pm 1.332)$ & $1.762(\pm 1.331)$& $1.763(\pm 1.331)$ &  $1.762(\pm 1.331)$\\
\hline
SFMEDM(D=8192) & $0.881(\pm 0.666)$ & $0.881(\pm 0.665)$ & $0.879(\pm 0.665)$ & $0.881(\pm 0.665)$ & $0.876(\pm 0.664)$\\
FMEDM(D=8192) & $0.880(\pm 0.666)$ & $0.881(\pm 0.666)$ & $0.881(\pm 0.666)$ & $0.881(\pm 0.666)$ &  $0.881(\pm 0.665)$\\
\hline
SFMEDM(D=16384) & $0.623(\pm 0.471)$ & $0.623(\pm 0.470)$ & $0.621(\pm 0.470)$ & $0.623(\pm 0.470)$ & $0.606(\pm 0.461)$\\
FMEDM(D=16384) & $0.628(\pm 0.470)$ & $0.623(\pm 0.471)$ & $0.623(\pm 0.471)$ & $0.623(\pm 0.471)$ & $0.623(\pm 0.471)$ \\
\hline
\end{tabular}
\label{tab:esperror}
\end{center}
\end{table*}
\begin{table*}[t]
\begin{center}
\caption{Average error by SFMCGK and FMCGK for each dataset. All values are multiplied by $10^2$.}
\vspace{-0.3cm}
\begin{tabular}{|l||c|c|c|c|c|}
\hline
{\bf Method}     & {\bf Protein} & {\bf DNA} & {\bf Music} & {\bf Sports} & {\bf Compound}\\
\hline\hline
SFMCGK(D=128) & $7.056(\pm 5.319)$ & $7.051(\pm 5.318)$ & $7.059(\pm 5.320)$ & $7.057(\pm 5.319)$ & $7.056(\pm 5.317)$ \\
FMCGK(D=128)  & $7.054(\pm 5.316)$ & $7.055(\pm 5.322)$ & $7.059(\pm 5.319)$ & $7.057(\pm 5.319)$ & $7.060(\pm 5.319)$ \\
\hline
SFMCGK(D=512) & $3.524(\pm 2.662)$ & $3.526(\pm 2.663)$ & $3.526(\pm 2.662)$ & $3.525(\pm 2.662)$ & $3.525(\pm 2.661)$ \\
FMCGK(D=512)  & $3.523(\pm 2.664)$ & $3.526(\pm 2.664)$ & $3.527(\pm 2.661)$ & $3.525(\pm 2.662)$ & $3.527(\pm 2.663)$ \\
\hline
SFMCGK(D=2048) & $1.761(\pm 1.331)$ & $1.762(\pm 1.332)$ & $1.763(\pm 1.332)$ & $1.763(\pm 1.332)$ & $1.762(\pm 1.331)$ \\
FMCGK(D=2048)  & $1.762(\pm 1.332)$ & $1.761(\pm 1.331)$ & $1.332(\pm 1.763)$ & $1.762(\pm 1.331)$ & $1.763(\pm 1.331)$\\
\hline 
SFMCGK(D=8192) & $0.881(\pm 0.662)$ & $0.881(\pm 0.665)$ & $0.881(\pm 0.665)$ & $0.881(\pm 0.665)$ & $0.869(\pm 0.663)$\\
FMCGK(D=8192) & $0.881(\pm 0.666)$  & $0.881(\pm 0.666)$ & $0.881(\pm 0.666)$ & $0.881(\pm 0.665)$ & $0.881(\pm 0.666)$\\
\hline
SFMCGK(D=16384) & $0.623(\pm 0.471)$ & $0.623(\pm 0.470)$ & $0.632(\pm 0.470)$ & $0.623(\pm 0.470)$ & $0.589(\pm 0.453)$\\
FMCGK(D=16384) & $0.623(\pm 0.471)$ & $0.623(\pm 0.470)$ & $0.632(\pm 0.471)$ & $0.623(\pm 0.470)$  & $0.623(\pm 0.470)$\\
\hline
\end{tabular}
\label{tab:cgkerror}
\end{center}
\end{table*}

\section{Feature Maps using CGK embedding}
CGK~\cite{Chakraborty16,Zhang17} is another string embedding using a randomized algorithm.
Let $S_i$ for $i=1$,$2$,$...$,$N$ be input strings of alphabet $\Sigma$ and
let $L$ be the maximum length of input strings.
CGK maps input strings $S_i$ in the edit distance space into strings $S^\prime_i$ of length $L$ in the Hamming space, i.e,
the edit distance between each pair $S_i$ and $S_j$ of input strings is approximately preserved by the Hamming distance of the corresponding pair $S^\prime_i$ and $S^\prime_j$
of the mapped strings. See \cite{Zhang17} for the detail of CGK.

To apply SFMs, we convert mapped strings $S^\prime_i$ in the Hamming space by CGK
to characteristic vectors $V^{C}(S^\prime_i)$ in the $L_1$ distance space as follows.
We view elements $S^\prime_i[j]$ for $j=1$,$2$,...,$L$ as locations (of the nonzero elements)
instead of characters.
For example, when $\Sigma=\{1,2,3\}$, we view each $S^\prime_i[j]$ as a vector of length $|\Sigma|=3$.
If $S^\prime_i[j]=1$, then we code it as $(0.5,0,0)$;
if  $S^\prime_i[j]=3$, then we code it as $(0,0,0.5)$.
We then concatenate those $L$ vectors into one vector $V^C(S^\prime_i)$ of dimension $L|\Sigma|$ and with $L$ nonzero elements. 
As a result, the Hamming distance between original strings $S^\prime_i$ and $S^\prime_j$ is equal to the $L_1$ distance between obtained 
vectors $V^C(S^\prime_i)$ and $V^C(S^\prime_j)$, i.e., 
$Ham(S^\prime_i,S^\prime_j)=||V^C(S_i^\prime)-V^C(S_j^\prime)||_1$.
By applying SFMs or FMs to $V^C(S^\prime_i)$, we built vectors of RFFs $\z(V^C(S^\prime_i))$. 
We shall call approximations of alignment kernels using CGK and SFMs (respectively, FMs) {\em SFMCGK} (respectively, {\em FMCGK}).

\section{Experiments}\label{sec:exp}
In this section, we evaluated the performance of SFMEDM with five massive string datasets, as shown in Table~\ref{tab:dataset}.
The "Protein" and "DNA" datasets consist of 3,238 human enzymes obtained from the KEGG GENES database~\cite{Kanehisa17}, respectively. 
Each enzyme in "DNA" was coded by a string consisting of four types of nucleotides or bases (i.e., A, T, G, and C). 
Similarity, each enzyme in "Protein" was coded by a string consisting of 20 types of amino acids. 
Enzymes belonging to the isomerases class in the enzyme commission~(EC) numbers in "DNA" and "Protein" have positive labels and the other enzymes have negative labels. 
The "Music" and "Sports" datasets consist of 10,261 and 296,337 reviews of musical instruments products and sports products in English from Amazon~\cite{He16, McAuley15}, respectively. 
Each review has a rating of five levels.
We assigned positive labels to reviews with four or five levels for rating and negative labels to the other reviews. 
The "Compound" dataset consists of 1,367,074 bioactive compounds obtained from the NCBI PubChem database~\cite{Kim16}. 
Each compound was coded by a string representation of chemical structures called {\em SMILES}. 
The biological activities of the compounds for human proteins were obtained from the ChEMBL database. 
In this study we focused on the biological activity for the human protein microtubule associated protein tau~(MAPT). 
The label of each compound corresponds to the presence or absence of biological activity for MAPT. 


All the methods were implemented by C++, and all the experiments were performed on one core of a quad-core Intel Xeon CPU E5-2680 (2.8GHz). 
The execution of each method was stopped if it did not finish within 48 hours in the experiments. 
Software and datasets used in this experiments are downloadable from \url{https://sites.google.com/view/alignmentkernels/home}.

\begin{table*}[t]
\begin{center}
\caption{Execution time in seconds for building feature vectors and computing kernel matrices in addition to training linear/non-linear SVM for each method.} 
\begin{tabular}{|l||r|r|r|r|r|}
\hline
{\bf Method} & {\bf Protein} & {\bf DNA} & {\bf Music} & {\bf Sports} & {\bf Compound} \\
\hline\hline
SFMEDM(D=128) & 5 & 8 & 11 & 204 & 261 \\
\hline
SFMEDM(D=512) & 22  &34 & 47 & 799 & 1,022 \\
\hline
SFMEDM(D=2048) & 93 & 138 & 193 & 3,149 & 4,101 \\
\hline
SFMEDM(D=8192) & 367 & 544 & 729 & 12,179 & 16,425 \\
\hline
SFMEDM(D=16384) & 725 & 1,081 & 1,430 & 24,282 & 32,651 \\
\hline\hline
SFMCGK(D=128) & 14 & 52 & 26 & 452 & 397 \\
\hline
SFMCGK(D=512) & 60 & 222 & 104 & 1,747 & 1,570 \\
\hline
SFMCGK(D=2048) & 237 & 981 & 415 & 7,156 & 6,252 \\
\hline
SFMCGK(D=8192) & 969 & 3,693 & 1,688 & 27,790 & 25,054 \\
\hline
SFMCGK(D=16384) & 1,937 & 7,596 & 3,366 & 53,482 & 49,060 \\
\hline\hline
D2KE(D=128) & 319 & 4,536 & 296 & 8,139 & 1,641 \\ 
\hline
D2KE(D=512) & 1,250 & 19,359 & 1244 & 34,827 & 6,869 \\
\hline
D2KE(D=2048) & 5,213 & 76,937 & 5,018 & 140,187 & 28,116 \\
\hline
D2KE(D=8192) & 21,208 & $>$48h & 19,716 & $>$48h & $>$48h \\
\hline
D2KE(D=16384) & 43,417 & $>$48h & 38,799 & $>$48h & $>$48h \\
\hline\hline
LAK &  31,718& - & -& -& - \\
\hline
GAK & 25,252 & $>$48h& 101,079& $>$48h& $>$48h \\
\hline
EDMKernel & 20 & 28 & 162 & $>$48h& $>$48h \\
\hline
STK17 & $3218$ & $917$ & $>$48h & $>$48h & $>$48h \\
\hline
\end{tabular}
\label{tab:learningtime}
\end{center}
\end{table*}

\subsection{Scalability of ESP}\label{sec:expesp}
First, we evaluated the scalability of ESP and CGK. 
Table~\ref{tab:espcgk} shows the execution time, memory in megabytes and dimension $d$ of characteristic vectors generated by ESP and CGK. 
ESP and CGK were practically fast enough to build characteristic vectors for large datasets. 
The executions of ESP and CGK finished within 60 seconds for "Compound" that was the largest dataset consisting of more than 1 million compounds. 
At most 1.5GB memory was consumed in the execution of ESP. 
These results demonstrated high scalability of ESP for massive datasets. 

For each dataset, characteristic vectors of very high dimensions were built by ESP and CGK. 
For example, 18 million dimension vectors were built by ESP for the "Sports" dataset. 
Applying the original FMs for RFFs to such high dimension characteristic vectors consumed huge amount of memory, 
deteriorating the scalability of FMs. 
The proposed SFMs can solve the scalability problem, which will be shown in 
the next subsection. 

\subsection{Efficiency of SFMs}\label{sec:expsfm}
We evaluated the efficiency of SFMs applied to characteristic vectors built from ESP, and we compared SFMs with FMs. 
We examined combinations of characteristic vectors and projected vectors of 
SFMEDM, FMEDM, SFMCGK and FMCGK.
The dimension $D$ of projected vectors of RFFs was examined for $D=\{128, 512,$ $2048, 8192, 16384\}$. 

Figure~\ref{fig:rffmemory} shows the amount of memory consumed in SFMs and FMs for characteristic vectors built by ESP and CGK 
for each dataset. 
According to the figure, a huge amount of memory was consumed by FMs for high dimension characteristic vectors and projected vectors. 
Around 1.1TB and 323GB of memory were consumed by FMEDM for $D=16,384$ for "Sports" and "Compound", respectively. 
Those huge amounts of memory made it impossible to build high-dimension vectors of RFFs.
The memory required by SFMs was linear in regard to dimension $d$ of characteristic vectors for each dataset. 
Only 280MB and 80MB of memory were consumed by SFMEDM for $D=16,384$ for "Sports" and "Compound", respectively.
These results suggest that compared with FMEDM, SFMEDM dramatically reduces the amount of required memory. 

Figure~\ref{fig:rfftime} shows the execution time for building projected vectors for each dataset. 
According to the figure, execution time increases linearly with dimension $D$ for each method and for "Compound", SFMs built 16,384-dimension vectors of RFFs in around nine hours. 

We evaluated accuracies of our approximations of alignment kernels in terms of
average error of RFFs, defined as
\[
\sum_{i=1}^N\sum_{j=i}^N |k(S_i,S_j) - \z(V(S_i))^\prime \z(V(S_j))|/(N(N+1)/2),
\]
where $k(S_i,S_j)$ is defined by Eq.~\ref{eq:kernel} and $\beta=1$ was fixed. 
Average error of SFMs was compared with that of FMs for each dataset. 
Table~\ref{tab:esperror} shows average error of SFMs and FMs using characteristic vectors built from ESP 
for each dataset. 
The average errors of SFMEDM and FMEDM are almost the same for all datasets and dimension $D$.
The accuracies of FMs were preserved in the case of SFMs, while the amount of memory required by FMs was dramatically reduced. 
The same tendencies were observed for average errors of SFMs in combination with CGK, as shown 
in Table~\ref{tab:cgkerror}.

\begin{figure*}[t]
  \begin{tabular}{cc}
    \begin{minipage}{0.3\textwidth}
      \centering
      \includegraphics[width=5cm]{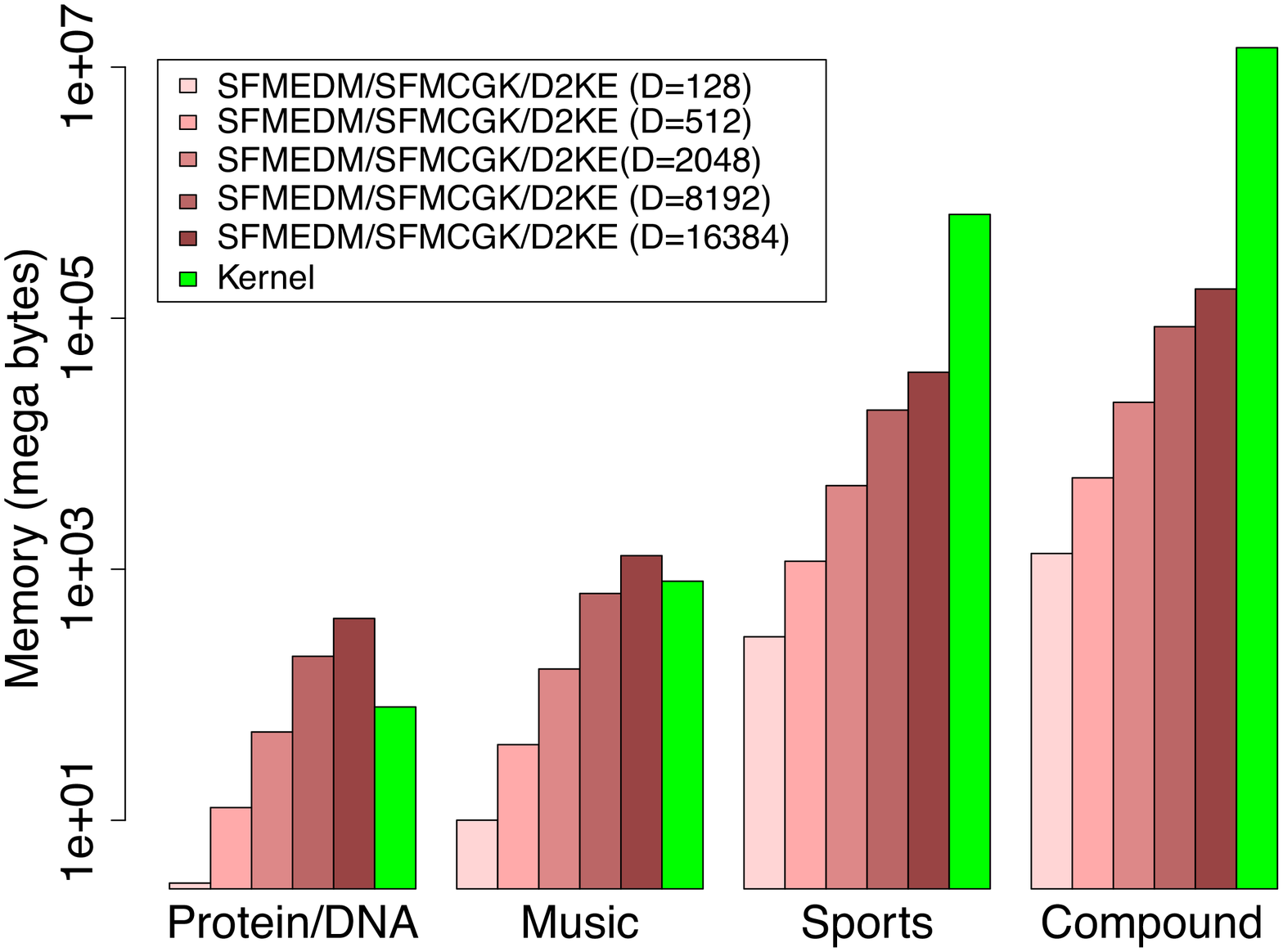}
      \vspace{-0.2cm}
      \caption{Memory in mega bytes for training SVM for each method. "Kernel" represents GAK, LAK, EDMKernel, CGKKernel and STK17.}
      \label{fig:learningmem}
      \end{minipage}
    \begin{minipage}{0.7\textwidth}
      \centering
      \includegraphics[width=11cm]{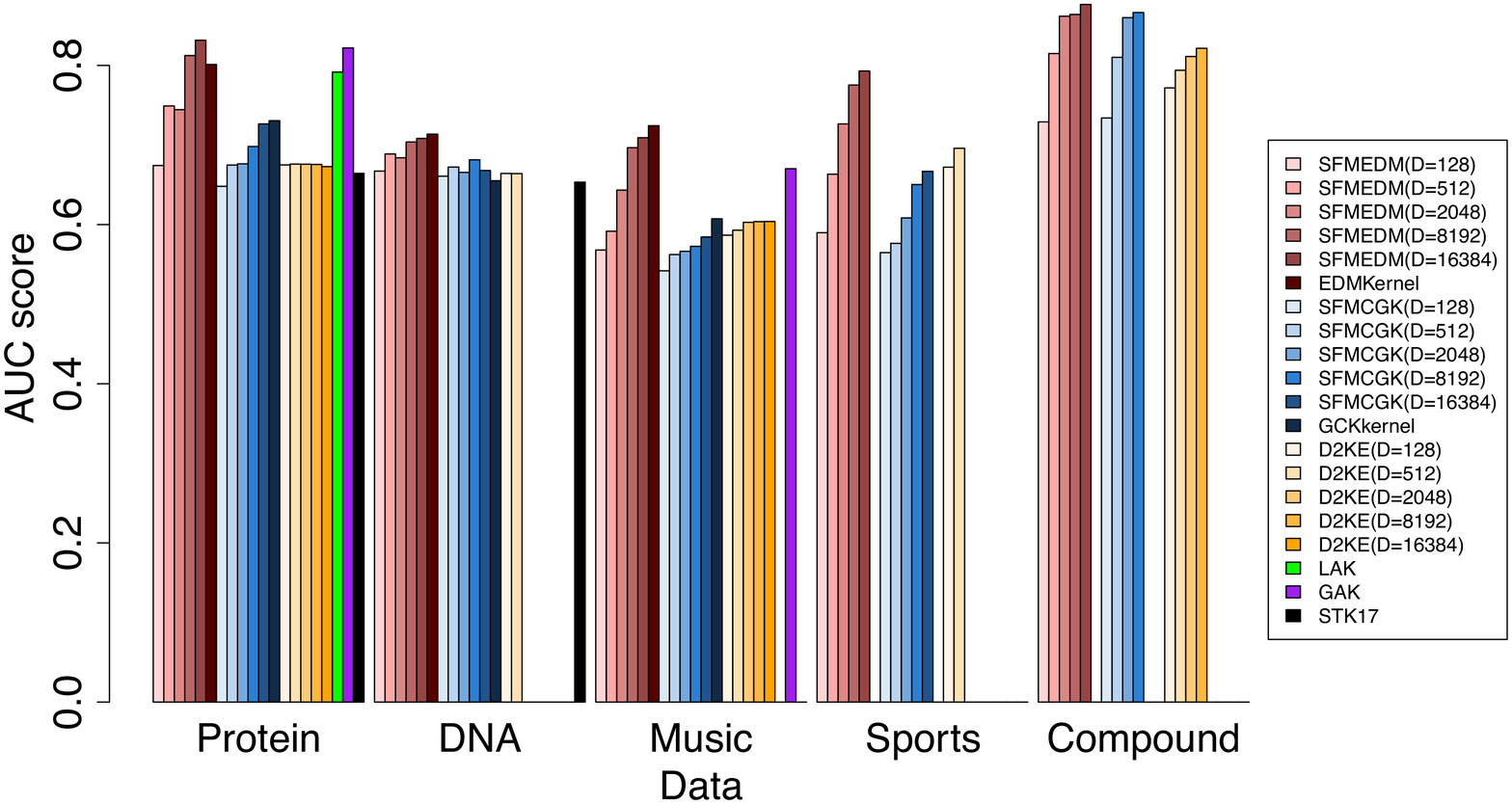}
      \vspace{-0.3cm}
      \caption{AUC score for each method.}
      \label{fig:auc}
      \end{minipage}
    \end{tabular}
\end{figure*}

\subsection{Classification performance of SFMEDM}\label{sec:expsvm}
We evaluated classification abilities of SFMEDM, SFMCGK, D2KE, LAK and GAK. 
We used an implementation of LAK downloadable from \url{http://sunflower.kuicr.kyoto-u.ac.jp/~hiroto/project/homology.html}.
We implemented D2KE by C++ with edit distance as a distance measure for strings. 
Laplacian kernels with characteristic vectors of ESP and CGK in Eq.\ref{eq:kernel} were also evaluated and denoted as 
ESPKernel and CGKKernel, respectively. 
In addition, we evaluated a classification ability of the state-of-the-art string kernel\cite{Farhan17}, which we shall refer to as {\em STK17}, and 
we used an implementation of STK17 downloadable from \url{https://github.com/mufarhan/sequence_class_NIPS_2017}.
We used LIBLINEAR~\cite{Fan08} for training linear SVM with SFMEDM and SFMCGK. 
We trained non-linear SVM with GAK, LAK, ESPKernel and CGKKernel using LIBSVM~\cite{Chang11}.
We performed three-fold cross-validation for each dataset and 
measured the prediction accuracy by {\em the area under the ROC curve (AUC)}.
Dimension $D$ of the vectors of RFFs and D2KE was examined for $D=\{128,$ $512,$ $2048,$ $8192,$ $16384\}$. 
We selected the best parameter achieving the highest AUC among all combinations of 
the kernel's parameter $\beta$$=$$\{1,$ $10,$ $100,$ $1000,$ $10000\}$ and the SVM's parameter $C=$$\{0.001,$ $0.01,$ $0.1,$ $1,$ $10,$ $100\}$.



Table~\ref{tab:learningtime} shows the execution time for building RFFs and computing kernel matrices in addition to training linear/non-linear SVM for each method. 
LAK was applied to only "Protein" because its scoring function was optimized for protein sequences.
It took 9 hours for LAK to finish the execution, which was the most time-consuming of all the methods in the case of "Protein". 
The execution of GAK finished within 48 hours for "Protein" and "Music" only, and it took around seven hours and 28 hours 
for "Protein" and "Music", respectively. 
The executions of D2KE did not finish within 48 hours for three large datasets of "Music", "Sports" and "Compound". 
In addition, the executions of EDMKernel and CGKKernel did not finish within 48 hours for "Sports" and "Compound". 
These results suggest that existing alignment kernels are unsuitable for applications to massive string datasets.
The executions of D2KE did not finish when large dimensions (e.g., $D=8,192$ and $D=16,384$) were used, which showed 
that creating high dimension vectors for achieving high classification accuracies by D2KE is time-consuming. 
The executions of SFMEDM and SFMCGK finished with 48 hours for all datasets.
SFMEDM and SFMCGK took around nine hours and 13 hours, respectively, for "Compound" consisting of 1.3-million strings in the setting of large $D=16,382$. 

Figure~\ref{fig:learningmem} shows amounts of memory consumed for training linear/non-linear SVM
for each method, where 
Here, GAK, LAK, EDMKernel, CGKKernel and STK17 are represented as "Kernel". 
"Kernel" required a small amount of memory for the small datasets (namely, "Protein", "DNA" and "Music"), 
but it required a huge amount of memory for the large datasets (namely, "Sports" and "Compound").
For example, it consumed 654 GB and 1.3 TB of memory for "Sports" and "Compound", respectively. 
The memories for SFMEDM, SFMCGK and D2KE were at least one order of magnitude smaller than those for "Kernel".
SFMEDM, SFMCGK and D2KE required 36GB and 166GB of memory for "Sports" and "Compound" in the case of large $D=16,382$, respectively.
These results demonstrated the high memory efficiency of SFMEDM and SFMCGK. 
Although training linear SVM with vectors built by D2KE was space-efficient, 
prediction accuracies were not high, which is presented next.

Figure~\ref{fig:auc} shows the classification accuracy of each method, where the results for the methods not finished with 48 hours were not plotted. 
The prediction accuracies of SFMEDM and SFMCGK were improved for larger $D$.
The prediction accuracy of SFMEDM was higher than that of SFMCGK for any $D$ on all datasets and was also higher than those of all the kernel methods (namely, LAK, GAK, ESPKernel and CGKKernel and STK17). The prediction accuracies of D2KE were worse than those of SFMEDM and were not improved for even large $D$. 
These results suggest that SFMEDM can achieve the highest classification accuracy and it is much more efficient than the other methods in terms of memory and time for building RFFs and training SVM. 



%

\section{Conclusion}
We have presented the first feature maps for alignment kernels, which we call SFMEDM, presented SFMs for computing RFFs space-efficiently, and demonstrated its ability to learn SVM for large-scale string classifications with various massive string data, and we demonstrate the superior performance of SFMEDM with respect to prediction accuracy, scalability and computation efficiency. 
Our SFMEDM has the following appealing properties: 
\begin{enumerate}
 \setlength{\parskip}{0cm}
 \setlength{\itemsep}{0cm}
\item {\bf Scalability: } SFMEDM is applicable to massive string data (see Section~\ref{sec:exp}).
\item {\bf Fast training: } SFMEDM trains SVMs fast (see Section~\ref{sec:expsvm}).
\item {\bf Space efficiency: } SFMEDM trains SVMs space-efficiently (see Section~\ref{sec:expsvm}).
\item {\bf Prediction accuracy: } SFMEDM can achieve high prediction accuracy (see Section~\ref{sec:expsvm}).
\end{enumerate}

SFMEDM opens the door to new application domains such as Bioinformatics 
and natural language processing, in which large-scale string processing with 
kernel methods was too restrictive so far. 

\section{Acknowledgments}
We thank Ninh Pham and Takaaki Nishimoto for useful discussions of kernel approximation and edit sensitive parsing. 
RP's work was supported by the European Research Council under the European Union’s 7th Framework Programme (FP7/2007-2013) / ERC grant agree-ment no.614331, and is also supported by Investigator Grant 16582, Basic Algorithms Research Copenhagen (BARC), from the VILLUM Foundation.
YY's work was supported by JSPS KAKENHI Grant Number 18H03334.

\bibliographystyle{IEEEtranS}
\bibliography{biblio}


\appendix

\section{Supplement}

\subsection{Edit sensitive parsing}

In the next section, we introduce {\em left preferential parsing~(LPP)} as a basic algorithm of ESP. 
In the later part of this section, we present the ESP algorithm.

\subsubsection{Left preferential parsing (LPP)}
The key idea of LPP is to make pairs of nodes from the left to the right positions preferentially in a sequence of nodes at an ESP tree and 
make triples of the remaining three nodes. 
Then, ESP builds type-2 nodes for these pairs of nodes and a type-1 node for the triple of nodes. 
In this way, LPP builds an ESP tree in a bottom-up manner. 

\begin{figure}
\begin{center}
\includegraphics[width=0.45\textwidth]{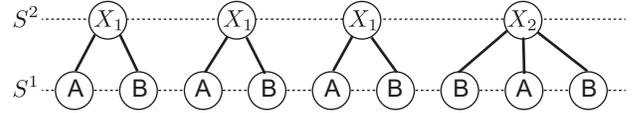}
\end{center}
\vspace{-0.3cm}
\caption{Example of LPP for a sequence $S^\prime=ABABABBAB$ with odd length}
\label{fig:lpp}
\end{figure}

More precisely, if the length of sequence $S^\ell$ at the $\ell$-th level of the ESP tree is even, 
LPP makes pairs of $S^\ell[2i-1]$ and $S^\ell[2i]$ for all $i \in [1,|S^\ell|/2]$ 
and builds type-2 nodes for all the pairs. 
Thus, $S^{\ell + 1}$ at the $(\ell+1)$-th level of the ESP tree is a sequence of type-2 nodes. 
If the length of sequence $S^{\ell + 1}$ of the $(\ell+1)$-th level of the ESP tree is odd, LPP makes pairs of $S^\ell[2i-1]$ and $S^\ell[2i]$ for each $i \in [1,|S^\ell/2 - 3|]$, and it makes triple of $S^{\ell}[2i-2]$, $S^{\ell}[2i-1]$ and $S^\ell[2i]$. 
LPP builds type-2 nodes for pairs of nodes and type-1 node for the triple of nodes. 
Thus, $S^{\ell+1}$ at the $(\ell+1)$-th level of the ESP tree is a sequence of type-2 nodes (except the last node) and a type-1 node as the last node. 
LPP builds an ESP tree in a bottom-up manner; that is, it build an ESP tree from leaves (i.e, $\ell = 1$) to the root. 
See Figure~\ref{fig:lpp} for an example of this ESP-tree building.

A crucial drawback of LPP is that it builds completely different ESP trees even for similar strings. 
For example, as shown in Figure~\ref{fig:exptree}, 
$S^\prime=AABABABBAB$ is a string where character $A$ is inserted at the first position of $S$.
Although $S^\prime$ and $S$ are similar strings, LPP builds completely different ESP trees, namely, $T^\prime$ and $T$ for $S^\prime$ and $S$, respectively, 
resulting in a large difference between EDM $EDM(S^\prime, S)$ and $L_1$ distance $||V(S^\prime) - V(S)||_1$ for characteristic vectors $V^\prime$ and $V$. 
Thus, LPP lacks the ability to approximate EDM.

\subsection{The ESP algorithm}
\begin{figure}[t]
\begin{center}
\includegraphics[width=0.45\textwidth]{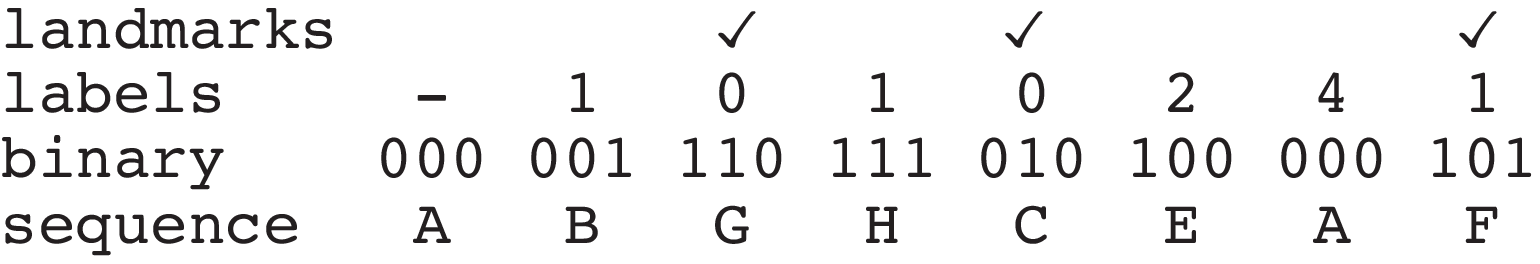}
\end{center}
\vspace{-0.7cm}
\caption{Example of alphabet reduction}
\label{fig:alphabetreduction}
\end{figure}  

ESP uses an engineered strategy while using LPP in its algorithm.
ESP classifies a string into substrings of three categories and applies different parsing strategies according to those categories. 
An ESP tree for an input string is built by gradually applying this parsing strategy of ESP to strings from the lowest to the highest level of the tree.

Given sequence $S^\ell$, ESP divides $S^\ell$ into subsequences in the following three categories: 
(i) a substring such that all pairs of adjacent node labels are different and substring length is at least $5$. 
Formally, a substring starting from position $s$ 
and ending at position $e$ in $S$ satisfies $S^\ell[i] \neq S^\ell[i+1]$ for any $i \in [s, e-1]$ and $(e - s + 1) \geq 5$;
(ii) a substring of the same node label and with length of at least $5$. 
Formally, a substring starting from position $s$ and ending at position $e$ satisfies 
$S^\ell[i] = S^\ell[i+1]$ for any $i \in [s, e-1]$ and $(e - s + 1) \geq 5$;
(iii) neither of categories (i) and (ii).

After classifying a sequence into subsequences of the above three categories, 
ESP applies different parsing methods to each substring according to their categories. 
ESP applies LPP to each subsequence of sequence $S^\ell$ in categories~(ii) and (iii), and it builds nodes at ($\ell+1$)-level.
For subsequences in category~(i), ESP applies a special parsing technique named {\em alphabet reduction}. 

\smallskip
{\em Alphabet reduction.} alphabet reduction is a procedure for converting a sequence to a new sequence with alphabet size of 3 at most. 
For each symbol $S^\ell[i]$, the conversion is performed as follows.
$S^\ell[i-1]$ is a left adjacent symbol of $S^\ell[i]$. Suppose $S^\ell[i-1]$ and $S^\ell[i]$ are represented as binary integers. 
Let $p$ be the index of the least-significant bit in which $S^\ell[i-1]$ differs from $S^\ell[i]$, and let $bit(p, S^\ell[i])$ 
be the binary integer of $S^\ell[i]$ at the $p$-th bit index. 
label $label(S^\ell[i])$ is defined as $2p + bit(p, S^\ell[i])$ and $label(S^\ell[i])$ is computed for each position $i$ in $S^\ell$.
When this conversion is applied to a sequence of alphabet $\Sigma$, the alphabet size of the resulting label sequence is $2\log{|\Sigma|}$, 
In addition, an important property of labels is that all adjacent labels in a label sequence are different, 
i.e., $label(S^\ell[i]) \neq label(S^\ell[i-1])$ for all $i \in [2,|S^\ell|]$. 
Thus, this conversion can be iteratively applied to a new label sequence, namely, $label(S^\ell[1])label(S^\ell[2])...label(S^\ell[L])$, until its alphabet size is at most $6$. 

The alphabet size is reduced from $\{0,1,...,5\}$ to $\{0,1,2\}$ as follows.
First, each $3$ in a sequence is replaced with the least element from $\{0,1,2\}$ that does not neighbor the $3$. 
Then, the same procedure is repeated for each $4$ and $5$, which generates a new sequence ($A$) of node labels drawn from $\{0,1,2\}$, 
where no adjacent characters are identical. 

Any position $i$ that is a {\em local maximum}, i.e., $A[i-1] < A[i] > A[i+1]$, is then selected.
Those positions are called {\em landmarks}.
In addition, any position $i$ that is a {\em local minimum}, i.e., $A[i-1] > A[i] < A[i+1]$, and 
not adjacent to an already chosen landmark, is selected as a landmark. 
An important property for those landmarks is that for any two successive landmark positions, $i$ and $j$,  either $|i-j|=2$ or $|i-j|=3$ hold, 
because $A$ is a sequence of no adjacent characters in alphabet $\{0,1,2\}$. 
Alphabet reduction for sequence $ABGHCEAF$ is illustrated in Figure~\ref{fig:alphabetreduction}.

Finally, type-2 nodes (respectively, type-3 nodes) are built for subsequences between landmarks $i$ and $j$ of length $|i-j|=2$ (respectively, $|i-j|=3$). 

The computation time of ESP is $O(NL)$.

\end{document}